\documentclass[sigconf]{acmart}

\AtBeginDocument{%
 }

\usepackage{algpseudocode}
\usepackage{algorithm}
\usepackage{balance}

\copyrightyear{2024}
\acmYear{2024}
\setcopyright{rightsretained}
\acmConference[KDD '24]{Proceedings of the 30th ACM SIGKDD Conference on Knowledge Discovery and Data Mining}{August 25--29, 2024}{Barcelona, Spain}
\acmBooktitle{Proceedings of the 30th ACM SIGKDD Conference on Knowledge Discovery and Data Mining (KDD '24), August 25--29, 2024, Barcelona, Spain}\acmDOI{10.1145/3637528.3672053}
\acmISBN{979-8-4007-0490-1/24/08}

\begin{document}

\title{Hierarchical Neural Constructive Solver for Real-world TSP Scenarios}

\author{Yong Liang Goh}
\affiliation{%
  \institution{Grabtaxi Holdings Pte Ltd \&\\ National University of Singapore}
  \country{Singapore}
}
\email{gyl@u.nus.edu}

\author{Zhiguang Cao}
\affiliation{%
  \institution{Singapore Management University}
  \country{Singapore}}
\email{zgcao@smu.edu.sg}

\author{Yining Ma}
\affiliation{%
  \institution{National University of Singapore}
  \country{Singapore}}
\email{yiningma@u.nus.edu}


\author{Yanfei Dong}
\affiliation{%
 \institution{National University of Singapore}
 \country{Singapore}}
\email{dyanfei@u.nus.edu}

\author{Mohammed Haroon Dupty}
\affiliation{%
  \institution{National University of Singapore}
  \country{Singapore}}
\email{haroon@nus.edu.sg}

\author{Wee Sun Lee}
\affiliation{%
  \institution{National University of Singapore}
  \country{Singapore}}
\email{leews@nus.edu.sg}



\renewcommand{\shortauthors}{Yong Liang Goh et al.}

\ccsdesc[500]{Computing methodologies~Machine learning}
\ccsdesc[500]{Computing methodologies~Neural networks}
\ccsdesc[500]{Computing methodologies~Sequential decision making}
\ccsdesc[500]{Computing methodologies~Learning latent representations}

\keywords{neural constructive solver, traveling salesman problem, deep reinforcement learning}



\label{sec:abstract}

\begin{abstract}
Existing neural constructive solvers for routing problems have predominantly employed transformer architectures, conceptualizing the route construction as a set-to-sequence learning task. However, their efficacy has primarily been demonstrated on entirely random problem instances that inadequately capture real-world scenarios. In this paper, we introduce realistic Traveling Salesman Problem (TSP) scenarios relevant to industrial settings and derive the following insights: (1) The optimal next node (or city) to visit often lies within proximity to the current node, suggesting the potential benefits of biasing choices based on current locations. (2) Effectively solving the TSP requires robust tracking of unvisited nodes and warrants succinct grouping strategies. Building upon these insights, we propose integrating a learnable choice layer inspired by Hypernetworks to prioritize choices based on the current location, and a learnable approximate clustering algorithm inspired by the Expectation-Maximization algorithm to facilitate grouping the unvisited cities. Together, these two contributions form a hierarchical approach towards solving the realistic TSP by considering both immediate local neighbourhoods and learning an intermediate set of node representations. Our hierarchical approach yields superior performance compared to both classical and recent transformer models, showcasing the efficacy of the key designs.
  
\end{abstract}
\maketitle
\section{Introduction}
\label{sec:intro}

The Traveling Salesman Problem (TSP) is a classical combinatorial optimization problem. Simply put, the TSP asks the following: given a set of cities, what is the shortest route where a salesman can visit every city only once and return back to his starting city? While it is simple to state, the TSP is a very difficult problem known to be NP-hard. Nevertheless, the TSP is a crucial problem to study, as many parallel problems can be reduced to solving the TSP, such as chip placement \cite{kumar2003optimizing}, the study of spin glass problems in physics \cite{kirkpatrick1985configuration}, DNA sequencing \cite{caserta2014hybrid}, and many others.

\begin{figure}
    \centering
    \includegraphics[width=1.0\linewidth]{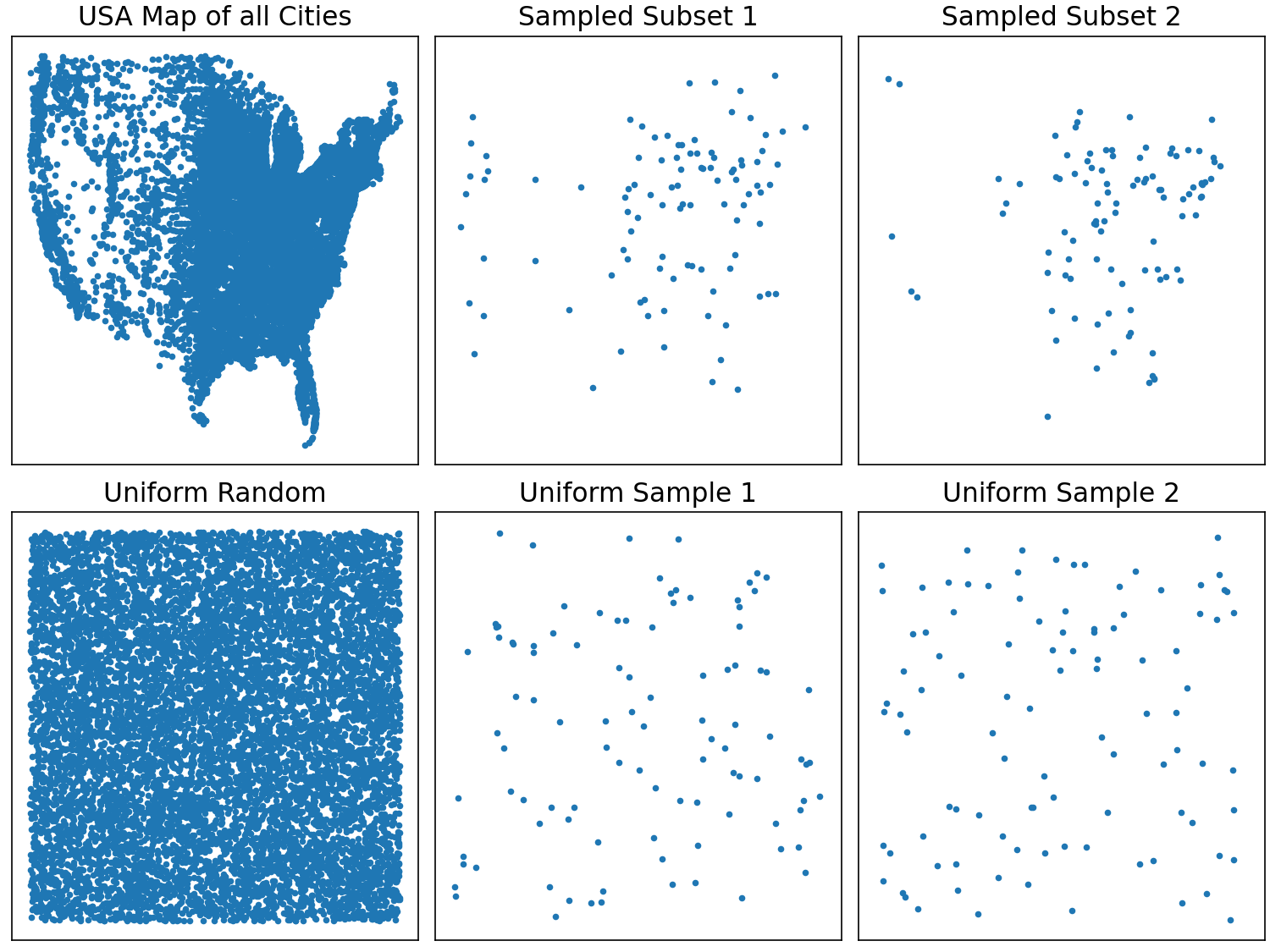}
    \caption{Comparing subset of TSP drawn from USA map (first row) against a uniform distribution (second row). Left-most plots show the base distribution. The subsets follow certain underlying structures for the USA case compared to completely random problems.}
    \label{fig:country_random}
\end{figure}

Given its prevalence across a multitude of domains, the TSP has been extensively researched in the community. Particularly, the main approaches can be broken down into exact methods and approximate methods. Exact methods often materialize in the form of mathematical programming. Some popular exact solvers, such as Concorde \cite{applegate2003concorde}, are developed based on linear programming and cutting planes. Approximate methods tend to be in the form of expert heuristics. An example would be the Lin-Kernighan-Helsgaun (LKH-3) algorithm, which utilizes heuristics and local search methods to update and improve initial solutions. As their names describe, exact methods return the true optimal routes while approximate ones return solutions often within some error bound of the optimal one. As the size of the problems grows, exact methods are intractable due to the NP-hard nature of the problem.

More recently, the deep learning community has put much effort into establishing practical neural solvers. These typically appear in the form of deep reinforcement learning \cite{bello2016neural, kool2018attention, kwon2020pomo, kim2022sym, gao2023towards}, which presents a label-free approach to improve the models. This is preferred over supervised learning approaches (e.g,~\cite{joshi2019efficient,sun2023difusco}) since they require large amounts of labelled data, which is often challenging to obtain given the limited scalability of exact solvers. 

It is important to note that learning-based solvers may perform well on specific target distributions they are trained on, but often suffer from poor generalization to other arbitrary instances. This is acceptable if the target distributions reflect real-world cases. However, prior works have mostly only been trained and tested on problems derived from random synthetic distributions, which may not accurately represent real-world applications. While some learned solvers have been tested on realistic TSP instances, such as those from the TSPLIB~\cite{Reinelt}, a collection of TSP instances observed in the real world, these realistic problem collections are typically too small to train on. Therefore, there is a gap in studying the performance of learning-based solvers in more realistic scenarios from the real world.

In this paper, we propose and study a setup that closely mirrors practical scenarios. Consider a logistic company with a large set of $M$ fixed locations that it can deliver to. The company will do many delivery trips to these locations but for each trip, only a small subset $V$ of the $M$ locations, e.g., locations that placed orders for that day, need to be visited. This observation motivates us to generate realistic distributions as follows: select a set of $M$ fixed locations from the real world to reflect real-world location distributions, and build each problem instance by randomly sampling a subset of $V$ locations from the set of $M$ locations.

We construct the locations using real-world datasets of cities from the USA, Japan, and Burma. When state-of-the-art neural solvers are trained in realistic settings, their performance may not be satisfactory. To improve the neural solvers, we focus on exploiting the nature of the node construction process. Existing neural solvers typically construct TSP tours autoregressively, solving the problem of visiting all unvisited cities starting from the current city and returning to the starting city. This suggests learning more effective and generalizable representations of the decision information regarding the current city and unvisited cities.

Specifically, we first observe that neural solvers often struggle to select a proper next city in the neighbourhood of the current city in our proposed realistic setup. This highlights the importance of exploiting more effective decision information about the current city. To this end, we propose to incorporate a customized hypernetwork layer \cite{ha2016hypernetworks} that leverages the embedding of the current city to modify the choice of the next city to visit.

Moreover, to obtain an improved representation of unvisited cities, we design a hierarchical representation that divides the cities into $C$ partitions and use an embedding to represent the unvisited cities in each of the $C$ partitions. We perform the partitioning using a soft clustering method, inspired by the EM algorithm. Given its differentiable property of EM, we can propagate the gradient through the clustering to learn the encoder parameters effectively. 

We showcase the effectiveness of the hypernetwork and the hierarchical representation of unvisited cities in experiments with the proposed realistic setting. We highlight the following contributions:
\begin{itemize}
    \item We introduce a more realistic TSP setting using real-world data to more convincingly demonstrate the effectiveness of neural TSP solvers.
    \item We make the key observation that neural solvers often struggle with node selection within a small locality, and we design a hypernetwork layer to emphasize local choices. 
    \item We further exploit the nature of solving structured TSPs by representing the set of unvisited cities with multiple key embeddings using a differentiable soft clustering algorithm instead of conventional simplistic pooling methods.
\end{itemize}

\section{Related Work}

\subsection{Constructive Neural Solvers} Deep reinforcement learning forms the hallmark of training constructive neural solvers. Early works from \cite{bello2016neural} proposed to use the Pointer Network \cite{vinyals2015pointer} based on the sequence-to-sequence architecture in \cite{sutskever2014sequence} to solve TSP and Knapsack problems. They employ an actor-critic approach and achieved strong results on the TSP. Follow-up works from \cite{nazari2018reinforcement} further improve the performance of the Pointer Network. 

Following on from this, the transformer network based on attention \cite{vaswani2017attention} was proposed by \cite{kool2018attention} to solve the TSP, Capacited Vehicle Routing Problem (CVRP) and the Orienteering Problem (OP). Primarily, the work showed that one can train a neural solver using the REINFORCE algorithm \cite{williams1992simple} and a simple greedy rollout of the network with a lagging baseline. Since then, multiple works based on the same architecture have been proposed to improve the predictive power of such solvers further~\cite{zhang2023review}. POMO \cite{kwon2020pomo} was introduced and observed that constructive solvers were limited by their starting nodes. Hence, to effectively explore the search space of solutions, one should use all nodes as starting nodes, effectively constructing a simple beam search. Additionally, they showed that a stable baseline can be found in the average of all solutions. Sym-NCO \cite{kim2022sym} was proposed to exploit the symmetry of TSP by introducing symmetry losses to regularize the transformer network. Recently, ELG was introduced by \cite{gao2023towards} that defined a local learnable policy based on a k-Nearest Neighbor graph. They exemplified the generalizability of the network on large CVRP instances. 

\subsection{Improvement Neural Solvers} Apart from constructive methods, another approach to solving the TSP looks at improvement solvers. This methodology is inspired by algorithms such as 2-opt, whereby the solver first starts with a complete route, and a heuristic is then used to select edges to delete or add so as to edit the solution. Such methods are measured based on how quickly they can reach a strong solution. Work such as \cite{d2020learning} uses neural networks to learn the 2-opt heuristic for improvement, while \cite{wu2021learning} uses the transformer to select node pairs for swaps for the TSP. Ma et al.~\cite{ma2021learning} then extended the transformer network to learn node and edge embeddings separately, which is then upgraded for pickup and delivery problems in \cite{ma2022efficient} and flexible k-opt in \cite{ma2023learning}, pushing the iterative solver's performance further.

\subsection{Search-based techniques} The previous two approaches are based on some form of learning-based search: constructive solvers try to perform a global search by learning heuristics entirely from data, whereas iterative solvers learn to guide local search techniques instead. 
Besides these, there is a class of search-based techniques that involve applying search during inference. Efficient Active Search (EAS) was proposed by \cite{hottung2021efficient} to introduce lightweight learnable layers at inference that could be tuned to improve the predictive power of a model on test samples. Other works such as \cite{fu2021generalize} showcase that one can leverage a small pre-trained network and combine it with search techniques such as Monte-Carlo Tree Search (MCTS) \cite{chaslot2008monte} so as to solve large-scale TSPs. The work in \cite{choo2022simulation} then combined both MCTS and EAS to improve the search capabilities further. Lastly, another work \cite{kool2022deep} showcased how one could combine dynamic programming with a pre-trained neural network to scale the TSP to 10,000 nodes.

Previous works attempt to regularize the networks via symmetry or scale the solver to larger problems. However, they essentially are still based on transformer models trained on arbitrary distributions. Apart from ELG, these works do not consider the impact of local choices nor explore further how to represent unvisited cities better, which are critical aspects of solving the TSP in our view.

\section{Our Approach}
\label{sec:app}


\begin{figure}
    \centering
    \includegraphics[width=1.0\linewidth]{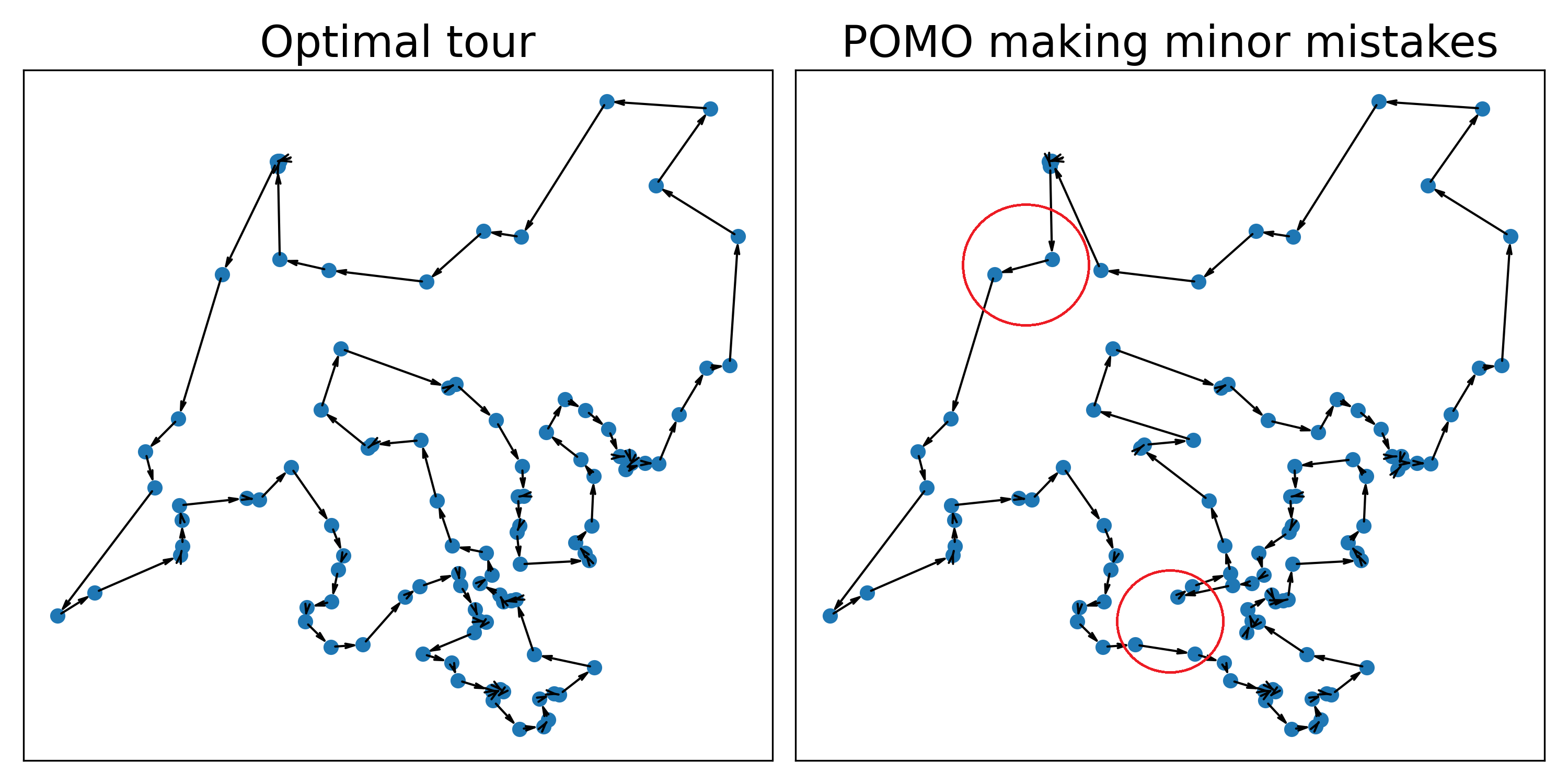}
    \caption{POMO model making a minor mistake with a poor selection of a local node}
    \label{fig:pomo_minor}
\end{figure}

\begin{figure}
    \centering
    \includegraphics[width=1.0\linewidth]{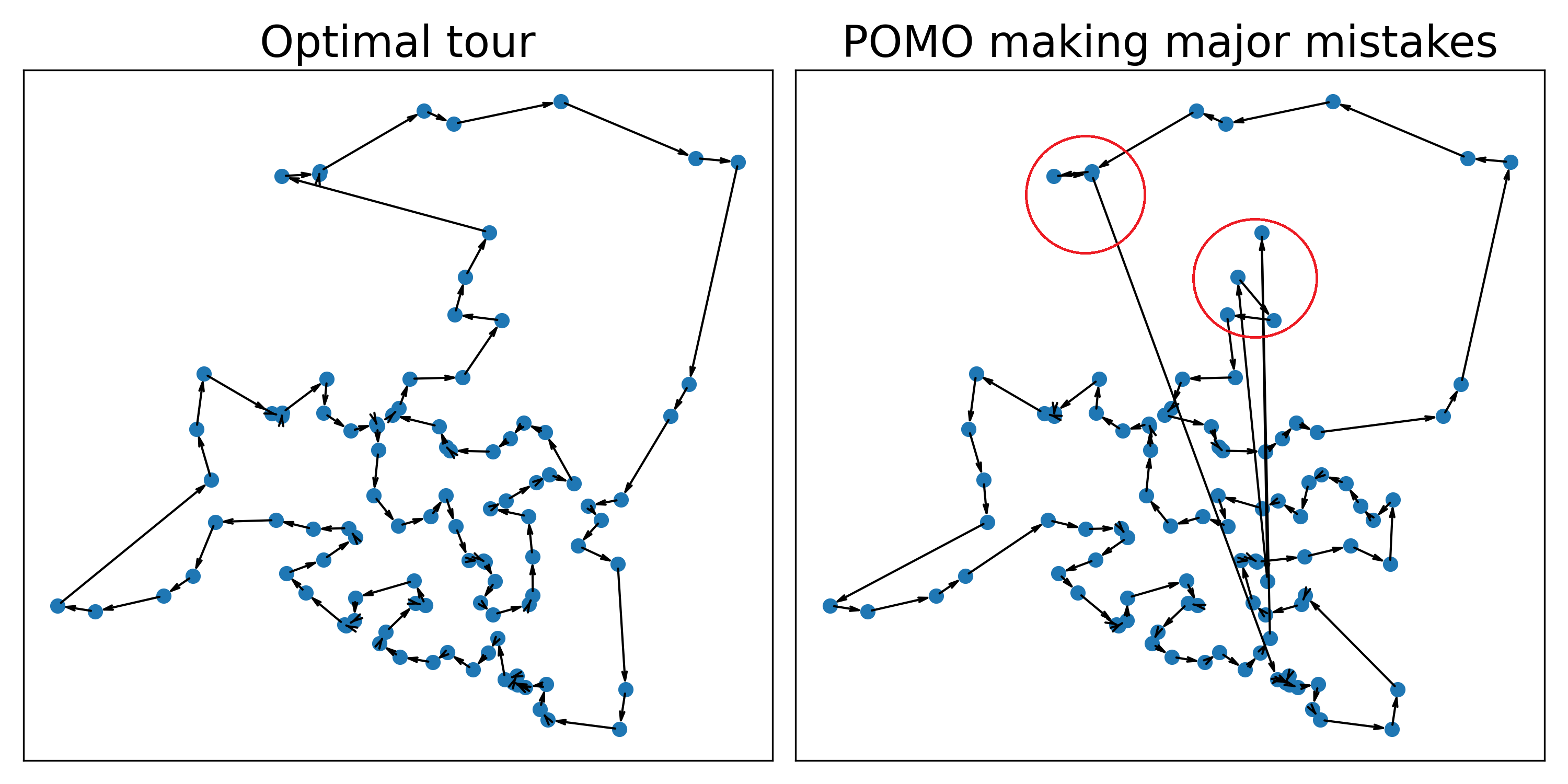}
    \caption{POMO model making a major mistake by not visiting nodes that are near it, causing cross-cluster routes that are inefficient}
    \label{fig:pomo_major}
\end{figure}

\begin{figure*}
    \centering
    \includegraphics[width=1.0\linewidth, height=12em]{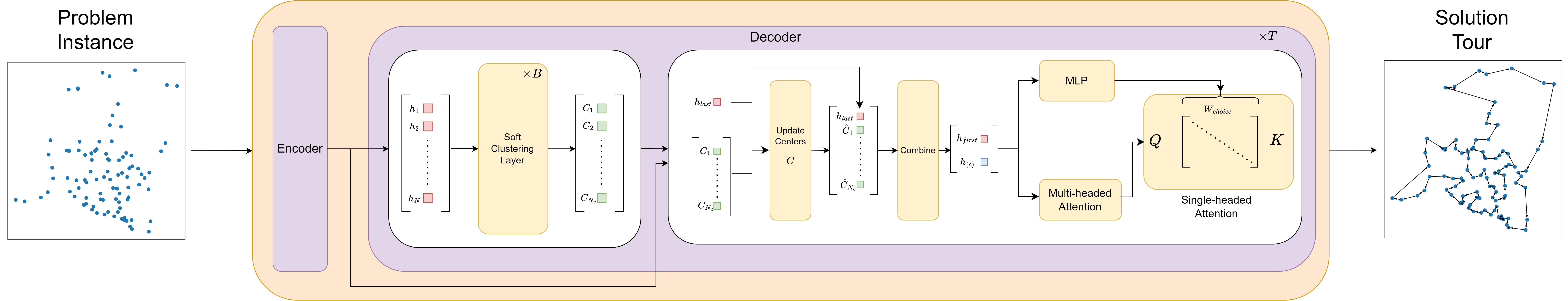}
    \caption{Overview of our proposed architecture. Given a TSP instance, we learn contextual embeddings of the cities in a set of cluster representations with an EM-inspired differentiable technique. In addition, our policy is dynamically adapted with a local hypernetwork which emphasises the completion of local cluster before moving on to new distant cities.}
    \label{fig:architecture}
\end{figure*}

Generally, we find that neural solvers tend to make two classes of errors in route construction compared to the optimal solution for such practical scenarios. The first class often appears as a minor error, where a poor decision is made in a local neighbourhood, as shown in Figure \ref{fig:pomo_minor}. This results in a sub-optimal route because a local choice is not picked first. The second class tends to appear in problems with more structure, where the agent fails to visit all reasonable nodes within a local cluster and has to backtrack to the area, exemplified in Figure \ref{fig:pomo_major}. This tends to give solutions that are significantly poorer than optimal.

Our approach seeks to tackle these two errors more effectively.
We propose two main architectural improvements to the base transformer model to address these issues. Firstly, we propose a learnable local choice decoder that accentuates certain choices based on the agent's current location (and hence locality). Secondly, we propose a differentiable clustering algorithm to learn a set of representations to capture and summarize the set of remaining cities. Our full approach is illustrated in Figure \ref{fig:architecture}.

\subsection{Recap: Constructive Neural Solvers}
In this subsection, we review previous works in well-known constructive neural solvers such as the attention model \cite{kool2018attention} and the POMO model \cite{kwon2020pomo}. The problem can be defined as an instance $s$ on a fully connected graph of $n$ nodes, where each node represents a city. Each node $n_i$ where $i \in \{1, ..., n\}$ has features $x_i$, typically the 2D coordinate. A solution is defined as a TSP tour, and is stated as a permutation of the nodes given by $\mathbf{\tau} = (\tau_1, ..., \tau_n)$, such that $\tau_t \in \{1, ..., n\}$. Note that since the tour does not allow backtracking, $\tau_t \neq \tau_{t'}, \forall t \neq t'$. The formulation yields the following policy:
\begin{equation}
    p_\theta (\mathbf{\tau} | s) = \prod^n_{t=1} p_\theta (\tau_t | s, \mathbf{\tau}_{1:t-1})
\end{equation}

Since the model is trained via reinforcement learning, the policy's action thus refers to the selection of the next node $\tau_t$ given the current state. The entire policy $p_\theta$ is parameterized with a neural network which involves both an encoder and a decoder. The encoder is a standard transformer model represented as such
\begin{equation}
    \mathbf{\tilde{h}}_i = \textsc{LN}^l ( \mathbf{h}_i^{l-1} + \textsc{MHA}_i^l(h_1^{l-1}, ..., h_n^{l-1})) 
\end{equation}
\begin{equation}
    \mathbf{h}_i^{l} = \textsc{LN}^l (\mathbf{\tilde{h}_i} + \textsc{FF}(\mathbf{\tilde{h}_i}))
\end{equation}
where $h^l_i$ is the embedding of the $i$-th node at the $l$-th layer, $d$ the dimension size of the embeddings, $\textsc{MHA}$ is the standard multi-headed attention layer \cite{vaswani2017attention}, $\textsc{LN}$ is a layer normalization function, and $\textsc{FF}$ is a simple feed-forward multi-layer perceptron (MLP). Each node's embeddings go through a total of $L$-layers before being passed into a decoder.

For the decoder, the solution is produced in an autoregressive fashion. In this case, at time step $t$, the decoder receives the following

\begin{equation}
    \mathbf{h}_{(c)} = \mathbf{h}_{\textsc{last}}^L + \mathbf{h}_{\textsc{start}}^L
\end{equation}
where $\mathbf{h}_{(c)}$ is known as a contextual embedding. In this instance, the context given is the sum of the $L$-th encoder layer's representation of the current node and the starting node. This is then first passed through a multi-headed attention layer where visited nodes are masked, followed by a single-headed attention layer for decision-making. In this decision layer, we obtain the following
\begin{equation}
    Q = W_Q(H), K = W_K(H), V = W_V(H)
\end{equation}
where $H$ is the set of all node embedding after the decoder's multi-headed attention layer. With this, we compute the compatibility of the query with all nodes together with a mask, where the mask indicates previously visited nodes. This ensures that the decoder cannot pick an already visited node. Mathematically, we use the following attention mechanism
\begin{equation}
\label{eqn:attn}
    a_j = 
    \begin{cases}
    U \cdot \textsc{tanh}(\frac{QK^\top}{\sqrt{d}}) & j \neq  \tau_{t'}, \forall t' < t\\
    -\infty & $\text{otherwise}$
    \end{cases}
\end{equation}
where we clip the values between $[-U, U]$ using a \textsc{tanh} function as with works in \cite{kool2018attention, bello2016neural, kwon2020pomo}. These values are then normalized with a simple softmax function, giving us the following decision-making policy:

\begin{equation}
\label{eqn:softmax}
    p_i = p_\theta(\tau_t = i|s, \tau_{1:t-1}) = \frac{e^{a_j}}{\sum_j e^{a_j}}
\end{equation}

Finally, to train the model, the REINFORCE algorithm \cite{williams1992simple} is used, where works such as POMO \cite{kwon2020pomo} and Sym-NCO \cite{kim2022sym} use a shared baseline of all starting points. Concretely, the expected return $J$ is maximized with gradient ascent and is approximated by

\begin{equation}
    \nabla_\theta J(\theta) \approx \frac{1}{N} \sum^{N}_{i=1} ( R(\tau^i) - b^i(s)) \nabla_\theta \log p_\theta (\tau^i | s)
\end{equation}
where $R(\tau^i)$ is the reward of permutation $\tau^i$, the tour length. $p_\theta(\tau^i|s)$ $ = \prod^T_{t=1} p_\theta(a^i_t|s, a^i_{1:t-1})$ is the product of action probabilities for the trajectory, where actions refer to the selection of the next node to move to, $s$ refers to the state of the problem, the set of all nodes, current node, and starting node.
The baseline is calculated as the average return of all starting points, given by
\begin{equation}
    b^i(s) = b_{\text{shared}} (s) = \frac{1}{N} \sum^{N}_{j=i} R(\tau^j) \forall i
\end{equation}

\subsection{Improving local decision making with the Choice Decoder}

More often than not, good selections for the TSP tend to lie within the locality of the current position. In the standard transformer architecture, the decoder attempts to capture this by using the current node's representation in a single-headed attention layer, as shown in Equations \ref{eqn:attn} and \ref{eqn:softmax}. In fact, Equation \ref{eqn:attn} is essentially a kernel function between the current querying node and the candidate nodes, an observation made in \cite{tsai2019transformer}.
Effectively, we can view the compatibility score as

\begin{equation}
    \phi(Q,K) = U \cdot \textsc{tanh}(\frac{QK^\top}{\sqrt{d}})
\end{equation}

Given that we aim to focus more on the current node's vicinity and features, we propose to \emph{generate a set of attention weights based on the current embeddings}. This aims to amplify or nullify the compatibility scores further by conditioning it on the current node. Hypernetworks \cite{ha2016hypernetworks} are small neural networks designed to generate a set of weights for a main network. Its goal is to serve as a form of relaxed weight sharing. This approach allows the hypernetwork to take as input some information about the problem, such as its structure, and adapt the main network's weights towards the problem. Inspired by this approach, we construct the set of attention weights using an MLP as a hypernetwork, with the input being the current node's embedding. Concretely, we modify the compatibility function as follows:

\begin{equation}
    \hat{a}_j = U \cdot \textsc{tanh}(\frac{(QW_{\textsc{choice}})K^\top}{\sqrt{d}})
\end{equation}
such that
\begin{equation}
    W_{\textsc{choice}} = \textsc{MLP}(Q), W_{\textsc{choice}} \in \mathrm{R}^{dxd}
\end{equation}

Effectively, $W_{\textsc{choice}}$ serves as a set of learnable conditional parameters that serve to alter the compatibility scores based on current embeddings. However, implementing this as a full matrix is extremely expensive. Instead, we realize $W_{\textsc{choice}}$ as a diagonal matrix, effectively reducing the compatibility score function to 
\begin{equation}
    \phi(Q,K|Q) = \tilde{a}_j = U \cdot \textsc{tanh}(\frac{(Q * W_{\textsc{choice}})K^\top}{\sqrt{d}})
\end{equation}
where $*$ is the element-wise product. Inherently, $W_{\textsc{choice}}$ now reduces to a set of learnable scalars which serve to modify the compatibility scores between $Q$ and $K$. Additionally, these learnable scalars are now conditioned upon $Q$, the current node, since it is generated via the MLP. The complexity of the MLP also now reduces from producing an output of $\mathrm{R}^{d \times d}$ to $\mathrm{R}^d$.

\subsection{Exploiting structure with Hierarchical Decoder}

In its current state, the decoder models the TSP as a set-to-sequence function. A key aspect of the input is the contextual embedding, $\mathbf{h}_{(c)}$. This embedding serves to represent the current state the model is in and is often a combination of the starting node, current node, and some global representation of the problem. For the global representation, works such as \cite{kool2018attention} and \cite{kwon2020pomo} use an average of all node embeddings, while others such as \cite{jin2023pointerformer}, maintain an average of all visited nodes so far. Essentially, all of these approaches attempt to capture various nuances of the TSP. 

However, for realistic problem settings, it is important to exploit the structure of the distribution of the cities. A single global representation would not be effective enough to capture the intricate correlations present between the cities. One notable and ubiquitous case is the presence of cluster patterns wherein certain cities are located near to each other while being distant to others. This clustering pattern, if captured within the global and contextual representation, can potentially provide the model with important clues to determine the next city to visit. Thus, for such problems, we propose to maintain a set of $C$ representations that are able to summarize the set of unvisited cities left, instead of a simple single representation. We postulate that this is meaningful as structured problems have frequent cities in fixed areas of the map, and being able to identify if a node belongs to a certain area could be beneficial to the decision-making process.

Prior works in other domains have shown the efficacy of cluster construction in applications such as node classification \cite{dong2024differentiable}. In this work, we wish to group all cities into $C$ representations. To this end,  we design the following layer inspired by the Expectation-Maximization (EM) algorithm \cite{moon1996expectation}. We first briefly review the EM algorithm for the Gaussian Mixture Model (GMM). Let $\mathbf{\theta} = \{\pi_k, \mathbf{\mu}_k\}$ denote the set of parameters, the coefficients of Gaussians $\pi$ and its associated means $\mu$ (covariance $\Sigma_k$ is assumed to be known), $\mathbf{X} = \{\mathbf{x}^{(i)}\}$ denote the set of data points, and $\mathbf{Z} = \{z^{(i)}\}$ denote the set of latent variables associated with the data.  The maximum-likelihood objective is given by
\begin{equation}
\label{eqn:em_loglikeli}
    \log p(\mathbf{X} | \pi, \mathbf{\mu}) = \sum^N_{i=1} \log \sum^K_{z^{(i)}=1} p(\mathbf{x}^{(n)} | z^{(n)}; \mathbf{\mu}) p(z^{(n)} | \pi)
\end{equation}
In general, Equation \ref{eqn:em_loglikeli} yields no closed-form solution. Additionally, it is non-convex, and its derivatives are expensive to compute. Since latent variable $z^{(i)}$ exists for every observation and we have a sum inside a log, we look at the EM algorithm to solve this. Typically, the EM algorithm involves two steps: the E-step computes the posterior probability over $z$ given the current parameters, and the M-step, which assumes that given that the data was generated with $z^{(i)} = k$, finds the set of parameters that maximizes this. Effectively, for a standard GMM, this yields the following E-step, given an initial set of parameters $\mathbf{\theta}$:
\begin{equation}
\label{eqn:em_estep}
    \gamma_k = p(z^{(i)} = k | \mathbf{x}^{(i)}; \mathbf{\theta}^{\text{old}}) = \frac{\pi_k \mathcal{N}(\mathbf{x}|\mathbf{\mu}_k)}{\sum^K_{j=1} \pi_j \mathcal{N}(\mathbf{x}|\mathbf{\mu}_j)}
\end{equation}
where $\gamma_k$ can be viewed as the responsibility of cluster $k$ towards data point $\mathbf{x}^{(i)}$. Then, for GMMs, the following update equations can be applied in the M-step:
\begin{equation}
\label{eqn:em_mstep1}
    \mathbf{\mu}_k = \frac{1}{N_k} \sum^N_{i=1} \gamma_k^{(i)} \mathbf{x}^{(i)}
\end{equation}


\begin{equation}
\label{eqn:em_mstep2}
    \pi_k = \frac{N_k}{N}, \text{where} \sum^N_{i=i} \gamma_k^{(i)}
\end{equation}

Essentially, Equation \ref{eqn:em_estep} estimates the contribution of each Gaussian model given the current set of parameters. While Equations \ref{eqn:em_mstep1} and \ref{eqn:em_mstep2}, highlight closed-form update equations for the Gaussian parameters.

Now, suppose a TSP instance drawn from a fixed map can be represented efficiently with $C$ latent representations. Since we are dealing with subsets of problems from the same space, these $C$ representations can be fixed and learnable. Let $\mathbf{C} \in \mathrm{R}^{N_c \times d}$ denote this set of representations, where we have $C = \{c_1, c_2, ..., c_{N_c} \}$. We propose to learn and update these representations by considering a mixture model, where the latent variables are modelled by these latent embeddings. Similar to the EM algorithm, we produce a set of mixing coefficients using an attention layer and its attention weights. Concretely, our soft clustering algorithm estimates its mixing coefficients via the attention mechanism, and using these scores, the clusters are then updated with a weighted sum of the embeddings. This can be shown as

\begin{equation}
    \hat{h}_i = W_H h_i
\end{equation}
\begin{equation}
    \hat{c_j} = W_C c_j
\end{equation}
\begin{equation}
\label{eqn:estep}
    \pi_{i,j} = \textsc{softmax}(\frac{\hat{h_i}\hat{c_j}^\top}{\sqrt{d}})
\end{equation}

\begin{equation}
\label{eqn:mstep}
    c_j = \sum_{i} \pi_{i,j} h_i
\end{equation}
where $\Pi$ is the matrix containing all coefficients $\pi_{i,j}$, $W_H$ and $W_C$ are parameters for the attentional scores, and $C$ is the set of learnable embeddings for the distribution. Given a single set of parameters in the attention layer, the embeddings $H$ and $C$ are passed through this same layer a total of $B$ times iteratively, mimicking a "rollout" of a soft clustering algorithm of E-steps and M-steps within each iteration. Loosely, we can see that Equation \ref{eqn:estep} resembles a similar calculation of the E-step, wherein we use a set of parameters to estimate the coefficients instead of minimizing for the Euclidean distance in the GMM case. Equation \ref{eqn:mstep} is similar to the M-step of GMMs, where we update the centers (in our case the embeddings are the latent variables) with a weighted sum of the data.

Once we have the set of $C$ embeddings, we update the representation at every step of decoding by subtracting a weighted sum of the current node's embedding; this computes the weighted sum of embeddings of unvisited cities, instead of all cities. Thus, at time step $T$, if the agent is current at node $i$, we update $C$ via
\begin{equation}
    c'_j = c_j - (\pi_{i,j} * h_i), \forall j \in N_c
\end{equation}

Then, we now construct a new context embedding such that
\begin{equation}
    \mathbf{h}_{(c)} =  W_{\textsc{combine}}[h^L_{\textsc{last}}, c_1, c_2, ..., c_{N_c}] + \mathbf{h}^L_{\textsc{first}}
\end{equation}
where $[\cdot]$ is the concatenation operation, and $W_\textsc{combine}$ is a simple linear layer to combine the embeddings. We keep $\mathbf{h}^L_{\textsc{first}}$ separate so as to preserve the importance of the starting node. As the decoder constructs the solutions, the $C$ embeddings get updated along the way, maintaining a small set of unvisited cities so as to keep track of the solution.

        

\begin{algorithm}
    \caption{Psuedo code of soft clustering algorithm}
    \label{alg:cluster}
    \begin{algorithmic}[1] 
        \Procedure{CLUSTER}{encoder embeddings $H$, number of centers $N_c$, number of iterations $B$, initial embeddings $C$, embedding size $d$}
            \For{$b \gets 1$ to $B$} 
                \State $\hat{H} \gets W_H(H)$
                \State $\hat{C} \gets W_C(C)$
                \State $\pi = \textsc{softmax}(\frac{\hat{H}\hat{C}^\top}{\sqrt{d}})$ \Comment{Compute attention scores}
                \State $C = \sum_i \pi_i h_i$ \Comment{Update the centers with data}
                \State $C_{\textsc{out}} = \hat{C} + C$ \Comment{Residual connection}
                \State $C = \textsc{Norm}(C_{\textsc{out}})$ \Comment{Layer normalization}
            \EndFor
        \State \textbf{return} $C$
        \EndProcedure
    \end{algorithmic}
\end{algorithm}

\begin{algorithm}
    \caption{Psuedo code of one step of decoding in the hierarchical neural constructive solver}
    \label{alg:decoder}
    \begin{algorithmic}[1] 
        \Function{UPDATE}{current node embedding $h_i$, cluster centers $C$, cluster weights $\pi$}
            \State $c'_j = c_j - (\pi_{i,j} * h_i), \forall c_j \in C$ 
            \State \textbf{return} $C'$
        \EndFunction
        
        \Procedure{DECODE}{encoder embeddings $H$, cluster centers $C$, cluster weights $\pi$, starting points $P$}
            \State $C' \gets \textsc{update}(h_{\textsc{last}}, C, \pi)$ \Comment{Remove visited embedding}
            \State $ \mathbf{h}_{(c)} \gets  W_{\textsc{combine}}[h_{\textsc{last}}, c_1, c_2, ..., c_{N_c}] + \mathbf{h}^L_{\textsc{first}}$
            \State $W_{choice} \gets \textsc{MLP}(h_{\textsc{last}})$ \Comment{Hypernetwork}
            \State $\mathbf{h'}_{(c)} \gets \textsc{MHA}( \mathbf{h}_{(c)}, K, V)$
            \State $\hat{a}_j = C \cdot \textsc{tanh}(\frac{(QW_{\textsc{choice}})K^\top}{\sqrt{d}})$
            \State $u_j \gets \textsc{softmax}(\hat{a}_j)$ \Comment{Create action probabilities}
            \State $p_i \gets \textsc{sample}(u)$ 
        \State \textbf{return} $i, p_i$\Comment{Return the selected node and its probability}
        \EndProcedure
    \end{algorithmic}
\end{algorithm}

In totality, our approach forms a hierarchy in the decision-making process; an intermediate level of $C$ embeddings represent the set of unvisited cities and their groupings, and immediate local decision-making is learnt and skewed by $W_\textsc{choice}$ to favour decisions based on current positions. Algorithms \ref{alg:cluster} and \ref{alg:decoder} highlights the overall flow of our approach.

\section{Experimental Setup}

\begin{table}[]
\caption{List of augmentations suggested by \cite{kwon2020pomo}}
\label{tbl:aug}
\begin{tabular}{@{}cc@{}}
\toprule
\multicolumn{2}{c}{$f(x,y)$} \\ \midrule
$(x,y)$       & $(y,x)$      \\
$(x, 1-y)$    & $(y, 1-x)$   \\
$(1-x, y)$    & $(1-y, x)$   \\
$(1-x, 1-y)$  & $(1-y, 1-x)$ \\ \bottomrule
\end{tabular}
\vspace{-10pt}
\end{table}
\subsection{Data Generation}

We present three different benchmarks for comparison. For all scenarios, we look at TSP-100 problems. Firstly, we generate random uniform data on a $[0,1]$ square and a fixed test set of $10,000$ instances. This is done to show if the addition of other layers interferes with the base performance of the transformer model.

\begin{figure}
    \centering
    \includegraphics[width=1.0\linewidth]{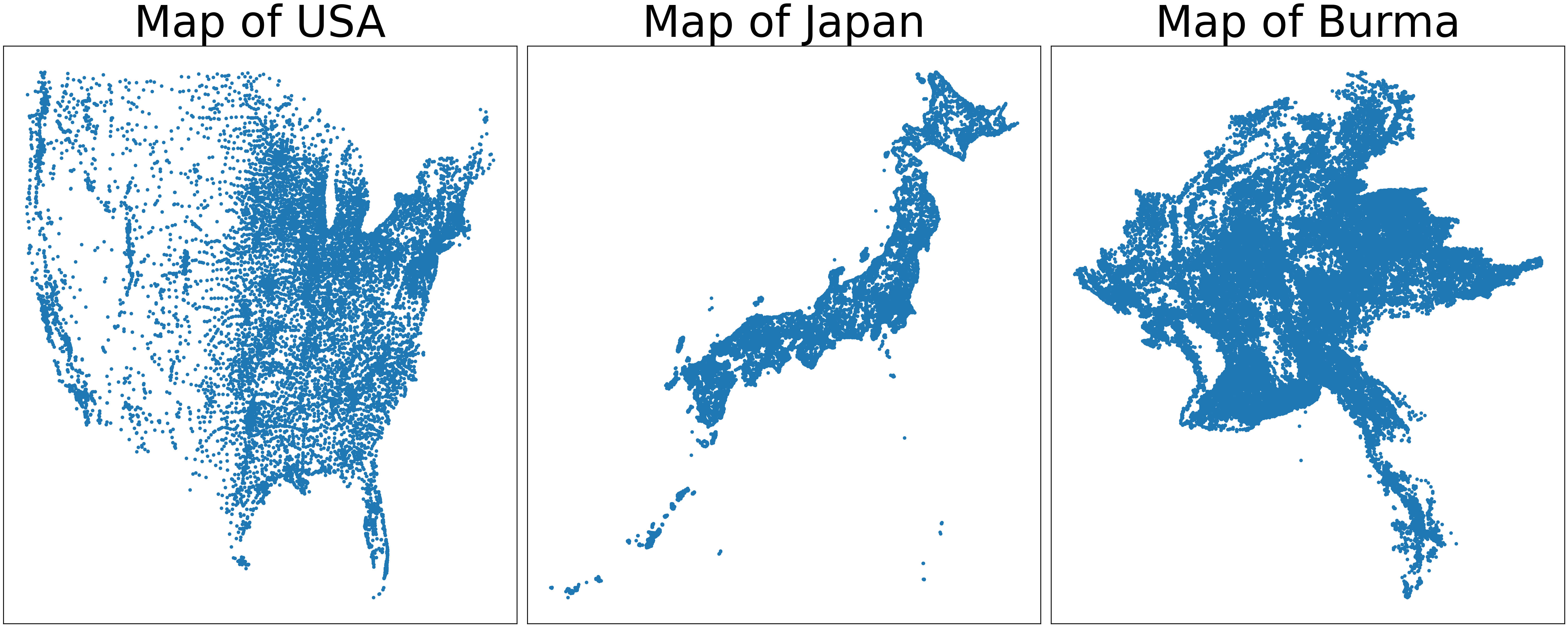}
    \caption{World maps used for experiments}
    \label{fig:all_world}
    \vspace{-10pt}
\end{figure}

To generate realistic data, we sample instances from 3 different countries, available at \cite{nattsp} and shown in Figure \ref{fig:all_world}. Namely, they are
\begin{itemize}
    \item USA13509 - 13,509 cities across the United States of America, each with a population >500
    \item BM33708 - 33,708 cities across the country of Burma
    \item JA9847 - 9,847 cities across the country of Japan
\end{itemize}

Each country is first normalized to a $[0,1]$ square. Then, at every training epoch, we randomly sample problems of size 100 from the map. Naturally, clusters that are denser across the country will be sampled more frequently. A test set size of $10,000$ samples is also drawn and held aside for evaluation. Each test set is fully solved via Concorde \cite{applegate2003concorde} to get the optimal length of each tour. We define 1 epoch to be $100,000$ samples, and the models are trained for $200$ epochs to prevent overfitting. In totality, the model sees $20,000,000$ different samples. Thirdly, we define a limited data setting. This mimics a typical practical problem where the company does not have unlimited access to data. We first define a small dataset of $50,000$ samples for such a setting. Based on the experiment size, we sample the necessary amount of data. Likewise, the models are tested on the same test set of $10,000$ samples.

Additionally, to show the generality of our approach beyond the logistics domain, we include the PCB3038 dataset from TSPLib, where the goal is to find the shortest path across a circuit board layout. Here, we can view the problem as such: from all possible holes the board can have, and given a subset of these holes, what is the optimal path of traversal? Solving such problems with high degrees of accuracy leads to cost savings in manufacturing and improved yields.


\begin{table*}[]
\caption{Model performance on 10,000 generated samples from the uniform random distribution. Best 
model in bold.}
\label{tbl:uniform}
\begin{tabular}{@{}ccccc@{}}
\toprule
Model                         & Tour Length     & Opt. Gap (\%)                          & Aug. Tour Length & Aug. Opt Gap (\%) \\ \midrule
\multicolumn{1}{c|}{Concorde} & 7.7649          & \multicolumn{1}{c|}{-}                 & 7.7649           & -                 \\
\multicolumn{1}{c|}{POMO}     & 7.8824          & \multicolumn{1}{c|}{1.5134\%}          & 7.8114           & 0.5997\%          \\
\multicolumn{1}{c|}{Sym-NCO}  & 7.9106          & \multicolumn{1}{c|}{1.8772\%}          & 7.8148           & 0.6425\%          \\
\multicolumn{1}{c|}{ELG}      & 7.8223          & \multicolumn{1}{c|}{0.7393\%}          & 7.7861           & 0.2734\%          \\
\multicolumn{1}{c|}{Ours}     & \textbf{7.8145} & \multicolumn{1}{c|}{\textbf{0.6397\%}} & \textbf{7.7809}  & \textbf{0.2063\%} \\ \bottomrule
\end{tabular}
\end{table*}

\begin{table*}[]
\caption{Performance of various models on realistic TSP100 instances from 3 different countries. Best 
models are in bold.}
\label{tbl:main}
\begin{tabular}{@{}ccccccc@{}}
\toprule
Dataset                       & Model                         & Tour Length     & Opt. Gap (\%)                          & Aug. Tour Length & Aug. Opt. Gap (\%)                        & Inference Time (10k samples) \\ \midrule
\multicolumn{1}{c|}{}         & \multicolumn{1}{c|}{Concorde} & 5.6209          & \multicolumn{1}{c|}{-}                 & 5.6209           & \multicolumn{1}{c|}{-}                    & 56 min 37 sec \\
\multicolumn{1}{c|}{}         & \multicolumn{1}{c|}{POMO}     & 5.6958          & \multicolumn{1}{c|}{1.3334\%}          & 5.6922           & \multicolumn{1}{c|}{1.2677\%}             & 2 min 46 sec \\
\multicolumn{1}{c|}{USA13509} & \multicolumn{1}{c|}{Sym-NCO}  & 5.7022          & \multicolumn{1}{c|}{1.4650\%}          & 5.6604           & \multicolumn{1}{c|}{0.7219\%}             & 2 min 49 sec \\
\multicolumn{1}{c|}{}         & \multicolumn{1}{c|}{ELG}      & 5.6660          & \multicolumn{1}{c|}{0.8022\%}          & 5.6641           & \multicolumn{1}{c|}{0.7691\%}             & 2 min 58 sec \\
\multicolumn{1}{c|}{}         & \multicolumn{1}{c|}{Ours}     & \textbf{5.6548} & \multicolumn{1}{c|}{\textbf{0.6024\%}} & \textbf{5.6533}  & \multicolumn{1}{c|}{\textbf{0.5762\%}}    & 3 min 03 sec \\ \midrule
\multicolumn{1}{c|}{}         & \multicolumn{1}{c|}{Concorde} & 3.5341          & \multicolumn{1}{c|}{-}                 & 3.5341           & \multicolumn{1}{c|}{-}                    & 52 min 55 sec \\
\multicolumn{1}{c|}{}         & \multicolumn{1}{c|}{POMO}     & 3.5621          & \multicolumn{1}{c|}{0.7926\%}          & 3.5620           & \multicolumn{1}{c|}{0.7893\%}             & 2 min 38 sec \\
\multicolumn{1}{c|}{JA9857}   & \multicolumn{1}{c|}{Sym-NCO}  & 3.5670          & \multicolumn{1}{c|}{0.9315\%}          & 3.5497           & \multicolumn{1}{c|}{0.4421\%}             & 2 min 40 sec \\
\multicolumn{1}{c|}{}         & \multicolumn{1}{c|}{ELG}      & 3.5576          & \multicolumn{1}{c|}{0.6659\%}          & 3.5574           & \multicolumn{1}{c|}{0.6596\%}             & 2 min 49 sec \\
\multicolumn{1}{c|}{}         & \multicolumn{1}{c|}{Ours}     & \textbf{3.5438} & \multicolumn{1}{c|}{\textbf{0.2741\%}} & \textbf{3.5435}  & \multicolumn{1}{c|}{\textbf{0.2670\%}}    & 2 min 51 sec \\ \midrule
\multicolumn{1}{c|}{}         & \multicolumn{1}{c|}{Concorde} & 5.0019          & \multicolumn{1}{c|}{-}                 & 5.0019           & \multicolumn{1}{c|}{-}                    & 59 min 22 sec \\
\multicolumn{1}{c|}{}         & \multicolumn{1}{c|}{POMO}     & 5.0823          & \multicolumn{1}{c|}{1.6060\%}          & 5.0746           & \multicolumn{1}{c|}{1.4540\%}             & 2 min 55 sec \\
\multicolumn{1}{c|}{BM33708}  & \multicolumn{1}{c|}{Sym-NCO}  & 5.0561          & \multicolumn{1}{c|}{1.0828\%}          & 5.0354           & \multicolumn{1}{c|}{0.6683\%}             & 2 min 58 sec \\
\multicolumn{1}{c|}{}         & \multicolumn{1}{c|}{ELG}      & 5.0587          & \multicolumn{1}{c|}{1.1343\%}          & 5.0528           & \multicolumn{1}{c|}{1.0162\%}             & 3 min 02 sec \\
\multicolumn{1}{c|}{}         & \multicolumn{1}{c|}{Ours}     & \textbf{5.0383} & \multicolumn{1}{c|}{\textbf{0.7261\%}} & \textbf{5.0328}  & \multicolumn{1}{c|}{\textbf{0.6166\%}}    & 3 min 06 sec \\ \bottomrule
\end{tabular}
\end{table*}

\begin{table*}[]
\caption{Ablation study on the USA13509 map}
\label{tbl:ablation}
\begin{tabular}{@{}cccccc@{}}
\toprule
Dataset                       & Model                                          & Tour Length & Opt. Gap (\%)                 & Aug. Tour Length & Aug. Opt. Gap (\%) \\ \midrule
\multicolumn{1}{c|}{}         & \multicolumn{1}{c|}{POMO only}                      & 5.6958      & \multicolumn{1}{c|}{1.3334\%} & 5.6922           & 1.2677\%           \\
\multicolumn{1}{c|}{}         & \multicolumn{1}{c|}{POMO + Choice}               & 5.6676      & \multicolumn{1}{c|}{0.8300\%} & 5.6659           & 0.7997\%           \\
\multicolumn{1}{c|}{USA13509}         & \multicolumn{1}{c|}{POMO + Choice Free}               & 5.7381      & \multicolumn{1}{c|}{2.1038\%} & 5.7311           & 1.9791\%           \\
\multicolumn{1}{c|}{} & \multicolumn{1}{c|}{POMO + Choice + Average Tracking} & 5.6648      & \multicolumn{1}{c|}{0.7807\%} & 5.6638           & 0.7633\%           \\
\multicolumn{1}{c|}{}         & \multicolumn{1}{c|}{POMO + Choice + Soft Clustering Tracking}      & \textbf{5.6548}      & \multicolumn{1}{c|}{\textbf{0.6024\%}} & \textbf{5.6533}           & \textbf{0.5762\%}           \\ \bottomrule
\end{tabular}
\end{table*}

\subsection{Benchmark Models}
We compare our approach to the following constructive neural solvers: POMO \cite{kwon2020pomo} the classical transformer that forms the basis for many follow-up works, Sym-NCO \cite{kim2022sym} a follow-up work from POMO that improves neural solvers by exploiting problem symmetry, and ELG \cite{gao2023towards} a recent work that also focuses on locality by defining a separate local policy based on its k-Nearest Neighbors. All models are trained based on the POMO shared baseline. It should be noted that in ELG, the authors introduced a different training algorithm. Since we wish to compare the efficacy of the architectural contributions, we train the ELG model using POMO. We reimplement all models and ran on the proposed dataset for our experimental results.

\subsection{Hyperparameters}
Since the neural models all share the same underlying backbone POMO transformer model, we retain the same set of hyperparameters across them to ensure the contributions are purely architectural. We utilize 6 layers of the transformer encoder and 2 layers of the transformer decoder. All models are trained for 200 epochs, with 100,000 episodes per epoch and a $1e^{-4}$ learning rate using the Adam optimizer \cite{kingma2014adam}. Gradient clipping is set to $[-10, 10]$ for all models. For ELG, we used their recommended 50-nearest neighbours. As for our approach, we set $N_c=5$ embeddings and $B=5$ iterations for the hierarchical approach by using a validation set of 1,000 samples for verification. Additional details for hyperparameters can be found in Appendix \ref{apn:params}.

\subsection{Performance Metric}
All models are measured by the optimality gap, the percentage gap between the neural model's tour length and the optimal tour length. This is given by
\begin{equation}
    O = (\frac{\frac{1}{N} \sum_i^N R_i}{\frac{1}{N} \sum_i^N L_i} - 1) * 100
\end{equation}
where $L_i$ is the tour length of test instance $i$ computed by Concorde.
Also, we perform instance augmentation, just like in POMO, which involves various translations and reflections across the $x$ and $y$ axes \cite{kwon2020pomo}, as shown in Table \ref{tbl:aug}.

\section{Results}

\begin{figure*}
    \centering
    \includegraphics[width=0.85\linewidth]{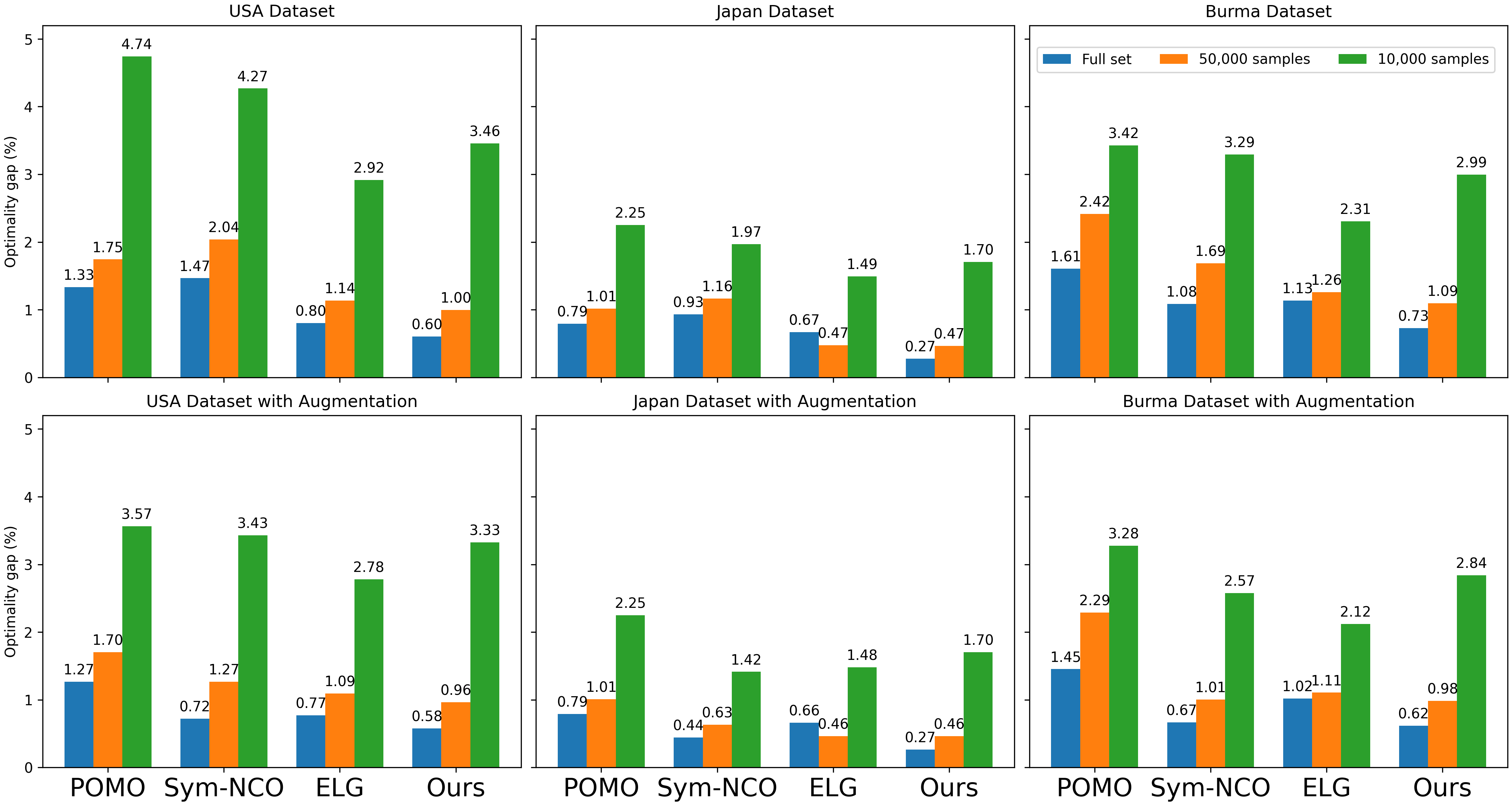}
    \caption{Optimality gaps across various dataset sizes (lower is better). The x-axis dictates the model types, and the y-axis denotes the optimality gap. The first row showcases standard data input, and the second shows data after augmentation. Countries are USA, Japan, and Burma, from left to right. Numerical details can be found in Appendix \ref{apn:bar_plot}.}
    \label{fig:multisize_dataset}
\end{figure*}

\subsection{Performance on Uniform Random Distribution}
Table \ref{tbl:uniform} highlights the overall performance of the models on the classic uniform random sampling dataset. Most models have solid performance, with augmentation playing a prominent role in boosting the predictive power. Our model and ELG also show the importance of having some form of local feature to improve the decision-making process. The addition of our unvisited city tracking via soft clustering provides additional boost to the model's predictive power since we can identify locations on the square.

\subsection{Performance on Structured Distributions}
Table \ref{tbl:main} showcases the different model's performance on TSP100 instances drawn from various countries. Our model has a clear advantage in the USA and Japan, with a narrow margin in Burma. Interestingly, we also see that augmentation has minimal effect on increasing the solver's performance, except for the case of Sym-NCO. It is likely because the maps are no longer symmetric in nature, and simple transformations do not improve the chances of finding a very different route. For Sym-NCO, we see that since it's specifically trained to exploit augmentations by considering all symmetries during training, its strong performance only appears when it can perform those transformations.
\subsection{Performance on Varied Sizes}
Figure \ref{fig:multisize_dataset} compares the model's performance between the complete set of training, 50,000 fixed samples, and 10,000 fixed samples. We can see that the model performance degrades as expected once data is limited. However, the ranking of the models is still similar at the 50,000 sample mark. Interestingly, reducing the dataset further to 10,000 samples sees ELG becoming the top neural constructive solver. We attribute ELG's strong performance on limited data to its local policy scheme. ELG uses well-crafted local features in the form of polar coordinates to create their local policy. The direct use of these features allows their local policy to learn valuable features to alter the action probabilities easily with less data. It should be noted that our choice hypernetwork is possibly a larger function class that also encompasses this approach. Hence, our model can learn a better overall function when more data is present.
\subsection{Performance on PCB3038}
Table \ref{tbl:pcb3038} showcases the performance of the models on the PCB3038 TSPLib dataset. Evidently, the dataset is no longer as symmetric as before, since Sym-NCO struggles to greatly improve upon POMO. Instead, some local features are important to have, as displayed by ELG and our model's improvement over POMO. Since our approach is a superset function of ELG, it can learn a better-performing solver, significantly improving upon POMO.

\subsection{Evolution of Embeddings}
\begin{figure}
    \includegraphics[width=0.88\linewidth]{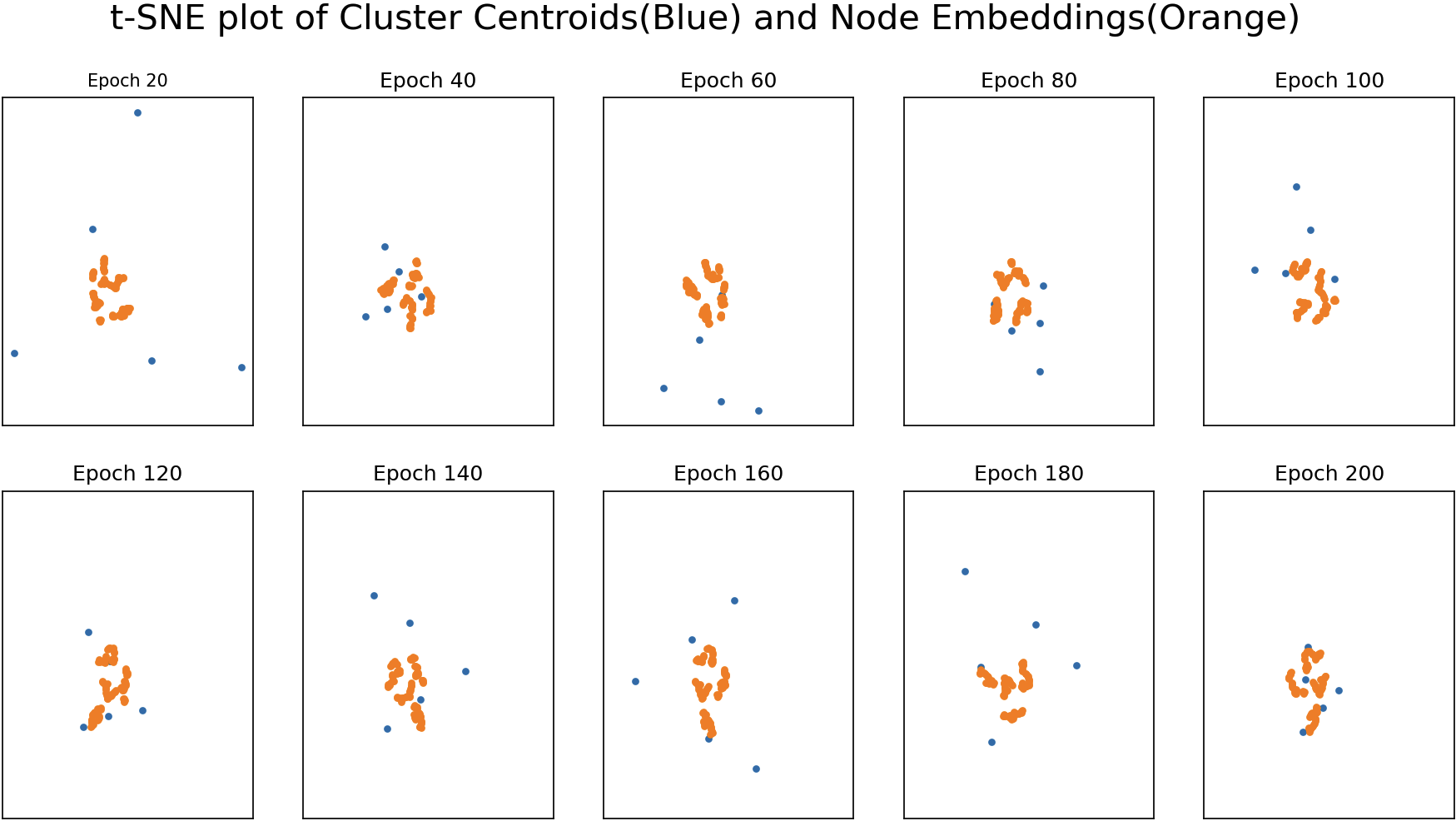}
    \caption{2D t-SNE plot of cluster centroids (in blue) and a set of node embeddings (in orange) as the training progresses. Over time, the centroids surround and segregate the embeddings better.}
    \label{fig:tsne}
\end{figure}

Soft clustering the nodes in the embedding space is a crucial part of our overall approach. Therefore, we evaluate if our algorithm indeed is able to cluster the nodes into a set of meaningful clusters as the training progresses. Specifically, we conduct a 2D t-SNE plot of the node and cluster embeddings during the training process. Figure \ref{fig:tsne} evidently shows that as the training progresses, the cluster centroid embeddings are better able to separate the node embeddings in the latent space. This plausibly leads to a better representation of the unvisited embeddings and hence the problem space, allowing the model to identify the groups of nodes so as to select similar ones first.

\begin{table*}[]
\caption{Model performance on 10,000 on the PCB3038 dataset to show efficacy of models on problems from domains beyond logistics.}
\label{tbl:pcb3038}
\begin{tabular}{@{}cccccc@{}}
\toprule
Dataset                      & Model                         & Tour Length     & Opt. Gap (\%)                          & Aug. Tour Length & Aug. Opt. Gap (\%) \\ \midrule
\multicolumn{1}{c|}{}        & \multicolumn{1}{c|}{Concorde} & 7.5866          & \multicolumn{1}{c|}{-}                 & 7.5866           & -                  \\
\multicolumn{1}{c|}{}        & \multicolumn{1}{c|}{POMO}     & 7.7952          & \multicolumn{1}{c|}{2.7494\%}          & 7.6984           & 1.4743\%           \\
\multicolumn{1}{c|}{PCB3038} & \multicolumn{1}{c|}{Sym-NCO}  & 7.7702          & \multicolumn{1}{c|}{2.4201\%}          & 7.6776           & 1.1996\%           \\
\multicolumn{1}{c|}{}        & \multicolumn{1}{c|}{ELG}      & 7.6882          & \multicolumn{1}{c|}{1.3393\%}          & 7.6387           & 0.6876\%           \\
\multicolumn{1}{c|}{}        & \multicolumn{1}{c|}{Ours}     & \textbf{7.6417} & \multicolumn{1}{c|}{\textbf{0.7236\%}} & \textbf{7.6074}  & \textbf{0.2746\%}  \\ \bottomrule
\end{tabular}
\end{table*}

\subsection{Ablation Studies}
As shown in Table \ref{tbl:ablation}, we perform ablation studies for our network and consider three different variations. Firstly, POMO + Choice adds only the local choice layer based on the hypernetwork. Secondly, POMO + Choice Free replaces the local choice hypernetwork layer with free parameters, meaning that the weights are no longer conditioned on the current node's embedding but rather allowed to learn freely. Thirdly, POMO + Choice + Average Tracking removes the soft clustering layer, instead averaging all unvisited nodes into a single embedding. From the table, we first see that the local choice layer is essential. Additionally, if we were to remove the ability to condition on the current node's position instead, the model would perform exceptionally poorly - even worse than the original transformer network. Finally, if we adopt the simplistic approach of averaging all unvisited cities, the model performance also suffers, showcasing the importance of our soft clustering layer.

\section{Conclusion}


In this work, we propose a more realistic approach to representing and generating Traveling Salesman Problems (TSPs) in real-world contexts. Our investigation reveals that previous state-of-the-art neural constructive solvers do not fully exploit the problem's intricacies to enhance predictive capability. To address this gap, we present a dual strategy to deal with the problem from two fronts. Firstly, we emphasize the importance of considering current agent positions, leading us to introduce a \emph{hypernetwork}, which enables dynamic fine-tuning to the decision-making process based on the agent's current node position. Secondly, we recognize that realistic TSP scenarios are often structured, and therefore, improving solutions in such scenarios necessitates a deeper understanding of the structure of the set of unvisited nodes. Instead of treating all unvisited nodes uniformly, we propose a \emph{soft clustering algorithm} inspired by the EM algorithm. This approach enhances the neural solver's performance by grouping nodes based on similarities, thereby increasing the likelihood of selecting nodes from the same cluster for early resolution. We illustrate the effectiveness of these methods across diverse geographical structures. Importantly, our methods are complementary and can be integrated with existing models like ELG or Sym-NCO to create more robust solutions.

\begin{acks}
This work was funded by the Grab-NUS AI Lab, a joint collaboration between GrabTaxi Holdings Pte. Ltd. and National University of Singapore, and the Industrial Postgraduate Program (Grant: S18-1198-IPP-II) funded by the Economic Development Board of Singapore. This research is also supported by the National Research Foundation, Singapore under its AI Singapore Programme (AISG Award No: AISG3-RP-2022-031).
\end{acks}

\bibliographystyle{ACM-Reference-Format}
\balance
\bibliography{biblo}


\begin{thebibliography}{36}


\ifx \showCODEN    \undefined \def \showCODEN     #1{\unskip}     \fi
\ifx \showDOI      \undefined \def \showDOI       #1{#1}\fi
\ifx \showISBNx    \undefined \def \showISBNx     #1{\unskip}     \fi
\ifx \showISBNxiii \undefined \def \showISBNxiii  #1{\unskip}     \fi
\ifx \showISSN     \undefined \def \showISSN      #1{\unskip}     \fi
\ifx \showLCCN     \undefined \def \showLCCN      #1{\unskip}     \fi
\ifx \shownote     \undefined \def \shownote      #1{#1}          \fi
\ifx \showarticletitle \undefined \def \showarticletitle #1{#1}   \fi
\ifx \showURL      \undefined \def \showURL       {\relax}        \fi
\providecommand\bibfield[2]{#2}
\providecommand\bibinfo[2]{#2}
\providecommand\natexlab[1]{#1}
\providecommand\showeprint[2][]{arXiv:#2}

\bibitem[nat({[n.\,d.]})]%
        {nattsp}
 \bibinfo{year}{[n.\,d.]}\natexlab{}.
\newblock
\newblock
\urldef\tempurl%
\url{https://www.math.uwaterloo.ca/tsp/world/countries.html}
\showURL{%
\tempurl}


\bibitem[Applegate(2003)]%
        {applegate2003concorde}
\bibfield{author}{\bibinfo{person}{David Applegate}.} \bibinfo{year}{2003}\natexlab{}.
\newblock \showarticletitle{Concorde: A code for solving traveling salesman problems}.
\newblock \bibinfo{journal}{\emph{http://www. tsp. gatech. edu/concorde. html}} (\bibinfo{year}{2003}).
\newblock


\bibitem[Bello et~al\mbox{.}(2016)]%
        {bello2016neural}
\bibfield{author}{\bibinfo{person}{Irwan Bello}, \bibinfo{person}{Hieu Pham}, \bibinfo{person}{Quoc~V Le}, \bibinfo{person}{Mohammad Norouzi}, {and} \bibinfo{person}{Samy Bengio}.} \bibinfo{year}{2016}\natexlab{}.
\newblock \showarticletitle{Neural combinatorial optimization with reinforcement learning}.
\newblock \bibinfo{journal}{\emph{arXiv preprint arXiv:1611.09940}} (\bibinfo{year}{2016}).
\newblock


\bibitem[Caserta and Vo{\ss}(2014)]%
        {caserta2014hybrid}
\bibfield{author}{\bibinfo{person}{Marco Caserta} {and} \bibinfo{person}{Stefan Vo{\ss}}.} \bibinfo{year}{2014}\natexlab{}.
\newblock \showarticletitle{A hybrid algorithm for the DNA sequencing problem}.
\newblock \bibinfo{journal}{\emph{Discrete Applied Mathematics}}  \bibinfo{volume}{163} (\bibinfo{year}{2014}), \bibinfo{pages}{87--99}.
\newblock


\bibitem[Chaslot et~al\mbox{.}(2008)]%
        {chaslot2008monte}
\bibfield{author}{\bibinfo{person}{Guillaume Chaslot}, \bibinfo{person}{Sander Bakkes}, \bibinfo{person}{Istvan Szita}, {and} \bibinfo{person}{Pieter Spronck}.} \bibinfo{year}{2008}\natexlab{}.
\newblock \showarticletitle{Monte-carlo tree search: A new framework for game ai}. In \bibinfo{booktitle}{\emph{Proceedings of the AAAI Conference on Artificial Intelligence and Interactive Digital Entertainment}}, Vol.~\bibinfo{volume}{4}. \bibinfo{pages}{216--217}.
\newblock


\bibitem[Choo et~al\mbox{.}(2022)]%
        {choo2022simulation}
\bibfield{author}{\bibinfo{person}{Jinho Choo}, \bibinfo{person}{Yeong-Dae Kwon}, \bibinfo{person}{Jihoon Kim}, \bibinfo{person}{Jeongwoo Jae}, \bibinfo{person}{Andr{\'e} Hottung}, \bibinfo{person}{Kevin Tierney}, {and} \bibinfo{person}{Youngjune Gwon}.} \bibinfo{year}{2022}\natexlab{}.
\newblock \showarticletitle{Simulation-guided beam search for neural combinatorial optimization}.
\newblock \bibinfo{journal}{\emph{Advances in Neural Information Processing Systems}}  \bibinfo{volume}{35} (\bibinfo{year}{2022}), \bibinfo{pages}{8760--8772}.
\newblock


\bibitem[d~O~Costa et~al\mbox{.}(2020)]%
        {d2020learning}
\bibfield{author}{\bibinfo{person}{Paulo~R d O~Costa}, \bibinfo{person}{Jason Rhuggenaath}, \bibinfo{person}{Yingqian Zhang}, {and} \bibinfo{person}{Alp Akcay}.} \bibinfo{year}{2020}\natexlab{}.
\newblock \showarticletitle{Learning 2-opt heuristics for the traveling salesman problem via deep reinforcement learning}. In \bibinfo{booktitle}{\emph{Asian conference on machine learning}}. PMLR, \bibinfo{pages}{465--480}.
\newblock


\bibitem[Dong et~al\mbox{.}(2024)]%
        {dong2024differentiable}
\bibfield{author}{\bibinfo{person}{Yanfei Dong}, \bibinfo{person}{Mohammed~Haroon Dupty}, \bibinfo{person}{Lambert Deng}, \bibinfo{person}{Zhuanghua Liu}, \bibinfo{person}{Yong~Liang Goh}, {and} \bibinfo{person}{Wee~Sun Lee}.} \bibinfo{year}{2024}\natexlab{}.
\newblock \bibinfo{title}{Differentiable Cluster Graph Neural Network}.
\newblock
\newblock
\showeprint[arxiv]{2405.16185}~[cs.LG]


\bibitem[Fu et~al\mbox{.}(2021)]%
        {fu2021generalize}
\bibfield{author}{\bibinfo{person}{Zhang-Hua Fu}, \bibinfo{person}{Kai-Bin Qiu}, {and} \bibinfo{person}{Hongyuan Zha}.} \bibinfo{year}{2021}\natexlab{}.
\newblock \showarticletitle{Generalize a small pre-trained model to arbitrarily large TSP instances}. In \bibinfo{booktitle}{\emph{Proceedings of the AAAI conference on artificial intelligence}}, Vol.~\bibinfo{volume}{35}. \bibinfo{pages}{7474--7482}.
\newblock


\bibitem[Gao et~al\mbox{.}(2023)]%
        {gao2023towards}
\bibfield{author}{\bibinfo{person}{Chengrui Gao}, \bibinfo{person}{Haopu Shang}, \bibinfo{person}{Ke Xue}, \bibinfo{person}{Dong Li}, {and} \bibinfo{person}{Chao Qian}.} \bibinfo{year}{2023}\natexlab{}.
\newblock \showarticletitle{Towards generalizable neural solvers for vehicle routing problems via ensemble with transferrable local policy}.
\newblock \bibinfo{journal}{\emph{arXiv preprint arXiv:2308.14104}} (\bibinfo{year}{2023}).
\newblock


\bibitem[Ha et~al\mbox{.}(2016)]%
        {ha2016hypernetworks}
\bibfield{author}{\bibinfo{person}{David Ha}, \bibinfo{person}{Andrew Dai}, {and} \bibinfo{person}{Quoc~V. Le}.} \bibinfo{year}{2016}\natexlab{}.
\newblock \bibinfo{title}{HyperNetworks}.
\newblock
\newblock
\showeprint[arxiv]{1609.09106}~[cs.LG]


\bibitem[Hottung et~al\mbox{.}(2021)]%
        {hottung2021efficient}
\bibfield{author}{\bibinfo{person}{Andr{\'e} Hottung}, \bibinfo{person}{Yeong-Dae Kwon}, {and} \bibinfo{person}{Kevin Tierney}.} \bibinfo{year}{2021}\natexlab{}.
\newblock \showarticletitle{Efficient active search for combinatorial optimization problems}.
\newblock \bibinfo{journal}{\emph{arXiv preprint arXiv:2106.05126}} (\bibinfo{year}{2021}).
\newblock


\bibitem[Jin et~al\mbox{.}(2023)]%
        {jin2023pointerformer}
\bibfield{author}{\bibinfo{person}{Yan Jin}, \bibinfo{person}{Yuandong Ding}, \bibinfo{person}{Xuanhao Pan}, \bibinfo{person}{Kun He}, \bibinfo{person}{Li Zhao}, \bibinfo{person}{Tao Qin}, \bibinfo{person}{Lei Song}, {and} \bibinfo{person}{Jiang Bian}.} \bibinfo{year}{2023}\natexlab{}.
\newblock \showarticletitle{Pointerformer: Deep Reinforced Multi-Pointer Transformer for the Traveling Salesman Problem}.
\newblock \bibinfo{journal}{\emph{arXiv preprint arXiv:2304.09407}} (\bibinfo{year}{2023}).
\newblock


\bibitem[Joshi et~al\mbox{.}(2019)]%
        {joshi2019efficient}
\bibfield{author}{\bibinfo{person}{Chaitanya~K Joshi}, \bibinfo{person}{Thomas Laurent}, {and} \bibinfo{person}{Xavier Bresson}.} \bibinfo{year}{2019}\natexlab{}.
\newblock \showarticletitle{An efficient graph convolutional network technique for the travelling salesman problem}.
\newblock \bibinfo{journal}{\emph{arXiv preprint arXiv:1906.01227}} (\bibinfo{year}{2019}).
\newblock


\bibitem[Kim et~al\mbox{.}(2022)]%
        {kim2022sym}
\bibfield{author}{\bibinfo{person}{Minsu Kim}, \bibinfo{person}{Junyoung Park}, {and} \bibinfo{person}{Jinkyoo Park}.} \bibinfo{year}{2022}\natexlab{}.
\newblock \showarticletitle{Sym-nco: Leveraging symmetricity for neural combinatorial optimization}.
\newblock \bibinfo{journal}{\emph{Advances in Neural Information Processing Systems}}  \bibinfo{volume}{35} (\bibinfo{year}{2022}), \bibinfo{pages}{1936--1949}.
\newblock


\bibitem[Kingma and Ba(2014)]%
        {kingma2014adam}
\bibfield{author}{\bibinfo{person}{Diederik~P Kingma} {and} \bibinfo{person}{Jimmy Ba}.} \bibinfo{year}{2014}\natexlab{}.
\newblock \showarticletitle{Adam: A method for stochastic optimization}.
\newblock \bibinfo{journal}{\emph{arXiv preprint arXiv:1412.6980}} (\bibinfo{year}{2014}).
\newblock


\bibitem[Kirkpatrick and Toulouse(1985)]%
        {kirkpatrick1985configuration}
\bibfield{author}{\bibinfo{person}{Scott Kirkpatrick} {and} \bibinfo{person}{G{\'e}rard Toulouse}.} \bibinfo{year}{1985}\natexlab{}.
\newblock \showarticletitle{Configuration space analysis of travelling salesman problems}.
\newblock \bibinfo{journal}{\emph{Journal de Physique}} \bibinfo{volume}{46}, \bibinfo{number}{8} (\bibinfo{year}{1985}), \bibinfo{pages}{1277--1292}.
\newblock


\bibitem[Kool et~al\mbox{.}(2022)]%
        {kool2022deep}
\bibfield{author}{\bibinfo{person}{Wouter Kool}, \bibinfo{person}{Herke van Hoof}, \bibinfo{person}{Joaquim Gromicho}, {and} \bibinfo{person}{Max Welling}.} \bibinfo{year}{2022}\natexlab{}.
\newblock \showarticletitle{Deep policy dynamic programming for vehicle routing problems}. In \bibinfo{booktitle}{\emph{International conference on integration of constraint programming, artificial intelligence, and operations research}}. Springer, \bibinfo{pages}{190--213}.
\newblock


\bibitem[Kool et~al\mbox{.}(2018)]%
        {kool2018attention}
\bibfield{author}{\bibinfo{person}{Wouter Kool}, \bibinfo{person}{Herke Van~Hoof}, {and} \bibinfo{person}{Max Welling}.} \bibinfo{year}{2018}\natexlab{}.
\newblock \showarticletitle{Attention, learn to solve routing problems!}
\newblock \bibinfo{journal}{\emph{arXiv preprint arXiv:1803.08475}} (\bibinfo{year}{2018}).
\newblock


\bibitem[Kumar and Luo(2003)]%
        {kumar2003optimizing}
\bibfield{author}{\bibinfo{person}{Ratnesh Kumar} {and} \bibinfo{person}{Zhonghui Luo}.} \bibinfo{year}{2003}\natexlab{}.
\newblock \showarticletitle{Optimizing the operation sequence of a chip placement machine using TSP model}.
\newblock \bibinfo{journal}{\emph{IEEE Transactions on Electronics Packaging Manufacturing}} \bibinfo{volume}{26}, \bibinfo{number}{1} (\bibinfo{year}{2003}), \bibinfo{pages}{14--21}.
\newblock


\bibitem[Kwon et~al\mbox{.}(2020)]%
        {kwon2020pomo}
\bibfield{author}{\bibinfo{person}{Yeong-Dae Kwon}, \bibinfo{person}{Jinho Choo}, \bibinfo{person}{Byoungjip Kim}, \bibinfo{person}{Iljoo Yoon}, \bibinfo{person}{Youngjune Gwon}, {and} \bibinfo{person}{Seungjai Min}.} \bibinfo{year}{2020}\natexlab{}.
\newblock \showarticletitle{Pomo: Policy optimization with multiple optima for reinforcement learning}.
\newblock \bibinfo{journal}{\emph{Advances in Neural Information Processing Systems}}  \bibinfo{volume}{33} (\bibinfo{year}{2020}), \bibinfo{pages}{21188--21198}.
\newblock


\bibitem[Ma et~al\mbox{.}(2023)]%
        {ma2023learning}
\bibfield{author}{\bibinfo{person}{Yining Ma}, \bibinfo{person}{Zhiguang Cao}, {and} \bibinfo{person}{Yeow~Meng Chee}.} \bibinfo{year}{2023}\natexlab{}.
\newblock \showarticletitle{Learning to Search Feasible and Infeasible Regions of Routing Problems with Flexible Neural k-Opt}. In \bibinfo{booktitle}{\emph{Thirty-seventh Conference on Neural Information Processing Systems}}.
\newblock


\bibitem[Ma et~al\mbox{.}(2022)]%
        {ma2022efficient}
\bibfield{author}{\bibinfo{person}{Yining Ma}, \bibinfo{person}{Jingwen Li}, \bibinfo{person}{Zhiguang Cao}, \bibinfo{person}{Wen Song}, \bibinfo{person}{Hongliang Guo}, \bibinfo{person}{Yuejiao Gong}, {and} \bibinfo{person}{Yeow~Meng Chee}.} \bibinfo{year}{2022}\natexlab{}.
\newblock \showarticletitle{Efficient Neural Neighborhood Search for Pickup and Delivery Problems}. In \bibinfo{booktitle}{\emph{Proceedings of the Thirty-First International Joint Conference on Artificial Intelligence}}. \bibinfo{pages}{4776--4784}.
\newblock


\bibitem[Ma et~al\mbox{.}(2021)]%
        {ma2021learning}
\bibfield{author}{\bibinfo{person}{Yining Ma}, \bibinfo{person}{Jingwen Li}, \bibinfo{person}{Zhiguang Cao}, \bibinfo{person}{Wen Song}, \bibinfo{person}{Le Zhang}, \bibinfo{person}{Zhenghua Chen}, {and} \bibinfo{person}{Jing Tang}.} \bibinfo{year}{2021}\natexlab{}.
\newblock \showarticletitle{Learning to iteratively solve routing problems with dual-aspect collaborative transformer}.
\newblock \bibinfo{journal}{\emph{Advances in Neural Information Processing Systems}}  \bibinfo{volume}{34} (\bibinfo{year}{2021}), \bibinfo{pages}{11096--11107}.
\newblock


\bibitem[Moon(1996)]%
        {moon1996expectation}
\bibfield{author}{\bibinfo{person}{Todd~K Moon}.} \bibinfo{year}{1996}\natexlab{}.
\newblock \showarticletitle{The expectation-maximization algorithm}.
\newblock \bibinfo{journal}{\emph{IEEE Signal processing magazine}} \bibinfo{volume}{13}, \bibinfo{number}{6} (\bibinfo{year}{1996}), \bibinfo{pages}{47--60}.
\newblock


\bibitem[Nazari et~al\mbox{.}(2018)]%
        {nazari2018reinforcement}
\bibfield{author}{\bibinfo{person}{Mohammadreza Nazari}, \bibinfo{person}{Afshin Oroojlooy}, \bibinfo{person}{Lawrence Snyder}, {and} \bibinfo{person}{Martin Tak{\'a}c}.} \bibinfo{year}{2018}\natexlab{}.
\newblock \showarticletitle{Reinforcement learning for solving the vehicle routing problem}.
\newblock \bibinfo{journal}{\emph{Advances in neural information processing systems}}  \bibinfo{volume}{31} (\bibinfo{year}{2018}).
\newblock


\bibitem[Reinelt({[n.\,d.]})]%
        {Reinelt}
\bibfield{author}{\bibinfo{person}{Gerhard Reinelt}.} \bibinfo{year}{[n.\,d.]}\natexlab{}.
\newblock
\newblock
\urldef\tempurl%
\url{http://comopt.ifi.uni-heidelberg.de/software/TSPLIB95/}
\showURL{%
\tempurl}


\bibitem[Sun and Yang(2023)]%
        {sun2023difusco}
\bibfield{author}{\bibinfo{person}{Zhiqing Sun} {and} \bibinfo{person}{Yiming Yang}.} \bibinfo{year}{2023}\natexlab{}.
\newblock \showarticletitle{Difusco: Graph-based diffusion solvers for combinatorial optimization}.
\newblock \bibinfo{journal}{\emph{arXiv preprint arXiv:2302.08224}} (\bibinfo{year}{2023}).
\newblock


\bibitem[Sutskever et~al\mbox{.}(2014)]%
        {sutskever2014sequence}
\bibfield{author}{\bibinfo{person}{Ilya Sutskever}, \bibinfo{person}{Oriol Vinyals}, {and} \bibinfo{person}{Quoc~V Le}.} \bibinfo{year}{2014}\natexlab{}.
\newblock \showarticletitle{Sequence to sequence learning with neural networks}.
\newblock \bibinfo{journal}{\emph{Advances in neural information processing systems}}  \bibinfo{volume}{27} (\bibinfo{year}{2014}).
\newblock


\bibitem[Tsai et~al\mbox{.}(2019)]%
        {tsai2019transformer}
\bibfield{author}{\bibinfo{person}{Yao-Hung~Hubert Tsai}, \bibinfo{person}{Shaojie Bai}, \bibinfo{person}{Makoto Yamada}, \bibinfo{person}{Louis-Philippe Morency}, {and} \bibinfo{person}{Ruslan Salakhutdinov}.} \bibinfo{year}{2019}\natexlab{}.
\newblock \showarticletitle{Transformer dissection: a unified understanding of transformer's attention via the lens of kernel}.
\newblock \bibinfo{journal}{\emph{arXiv preprint arXiv:1908.11775}} (\bibinfo{year}{2019}).
\newblock


\bibitem[Vaswani et~al\mbox{.}(2017)]%
        {vaswani2017attention}
\bibfield{author}{\bibinfo{person}{Ashish Vaswani}, \bibinfo{person}{Noam Shazeer}, \bibinfo{person}{Niki Parmar}, \bibinfo{person}{Jakob Uszkoreit}, \bibinfo{person}{Llion Jones}, \bibinfo{person}{Aidan~N Gomez}, \bibinfo{person}{{\L}ukasz Kaiser}, {and} \bibinfo{person}{Illia Polosukhin}.} \bibinfo{year}{2017}\natexlab{}.
\newblock \showarticletitle{Attention is all you need}.
\newblock \bibinfo{journal}{\emph{Advances in neural information processing systems}}  \bibinfo{volume}{30} (\bibinfo{year}{2017}).
\newblock


\bibitem[Vinyals et~al\mbox{.}(2015)]%
        {vinyals2015pointer}
\bibfield{author}{\bibinfo{person}{Oriol Vinyals}, \bibinfo{person}{Meire Fortunato}, {and} \bibinfo{person}{Navdeep Jaitly}.} \bibinfo{year}{2015}\natexlab{}.
\newblock \showarticletitle{Pointer networks}.
\newblock \bibinfo{journal}{\emph{Advances in neural information processing systems}}  \bibinfo{volume}{28} (\bibinfo{year}{2015}).
\newblock


\bibitem[Williams(1992)]%
        {williams1992simple}
\bibfield{author}{\bibinfo{person}{Ronald~J Williams}.} \bibinfo{year}{1992}\natexlab{}.
\newblock \showarticletitle{Simple statistical gradient-following algorithms for connectionist reinforcement learning}.
\newblock \bibinfo{journal}{\emph{Machine learning}}  \bibinfo{volume}{8} (\bibinfo{year}{1992}), \bibinfo{pages}{229--256}.
\newblock


\bibitem[Wu et~al\mbox{.}(2021)]%
        {wu2021learning}
\bibfield{author}{\bibinfo{person}{Yaoxin Wu}, \bibinfo{person}{Wen Song}, \bibinfo{person}{Zhiguang Cao}, \bibinfo{person}{Jie Zhang}, {and} \bibinfo{person}{Andrew Lim}.} \bibinfo{year}{2021}\natexlab{}.
\newblock \showarticletitle{Learning improvement heuristics for solving routing problems}.
\newblock \bibinfo{journal}{\emph{IEEE transactions on neural networks and learning systems}} \bibinfo{volume}{33}, \bibinfo{number}{9} (\bibinfo{year}{2021}), \bibinfo{pages}{5057--5069}.
\newblock


\bibitem[Ye et~al\mbox{.}(2024)]%
        {ye2024glop}
\bibfield{author}{\bibinfo{person}{Haoran Ye}, \bibinfo{person}{Jiarui Wang}, \bibinfo{person}{Helan Liang}, \bibinfo{person}{Zhiguang Cao}, \bibinfo{person}{Yong Li}, {and} \bibinfo{person}{Fanzhang Li}.} \bibinfo{year}{2024}\natexlab{}.
\newblock \showarticletitle{Glop: Learning global partition and local construction for solving large-scale routing problems in real-time}. In \bibinfo{booktitle}{\emph{Proceedings of the AAAI Conference on Artificial Intelligence}}, Vol.~\bibinfo{volume}{38}. \bibinfo{pages}{20284--20292}.
\newblock


\bibitem[Zhang et~al\mbox{.}(2023)]%
        {zhang2023review}
\bibfield{author}{\bibinfo{person}{Cong Zhang}, \bibinfo{person}{Yaoxin Wu}, \bibinfo{person}{Yining Ma}, \bibinfo{person}{Wen Song}, \bibinfo{person}{Zhang Le}, \bibinfo{person}{Zhiguang Cao}, {and} \bibinfo{person}{Jie Zhang}.} \bibinfo{year}{2023}\natexlab{}.
\newblock \showarticletitle{A review on learning to solve combinatorial optimisation problems in manufacturing}.
\newblock \bibinfo{journal}{\emph{IET Collaborative Intelligent Manufacturing}} \bibinfo{volume}{5}, \bibinfo{number}{1} (\bibinfo{year}{2023}), \bibinfo{pages}{e12072}.
\newblock


\end{thebibliography}

\appendix
\clearpage
\section{Appendix}
\subsection{Hyperparameters}\label{apn:params}
As described in the main body, we retain all original settings of the transformer. This yields the following:
\begin{itemize}
    \item 6 encoder layers
    \item 2 decoder layers
    \item We removed global graph embedding as it was found to harm performance
    \item Clipping value $U$ is set to $10.0$
    \item Batch size for training is $128$ samples
    \item A total of 200 epochs were run, where each epoch consisted of $100,000$ episodes
    \item Adam optimizer was used with a weight decay of $1e-6$
    \item Learning rate of $0.0001$ was used
    \item Unique to our model:
        \begin{itemize}
            \item The soft clustering layer is applied $B = 5$ times
            \item We use a total of $N_c = 5$ embeddings for the clustering
        \end{itemize}
\end{itemize}

\subsection{Model performance across varying dataset size}\label{apn:bar_plot}
Table \ref{tbl:bar_plot} showcases in detail the bar plot results from Figure \ref{fig:multisize_dataset}. Here, we compare the model's performance across three different sizes: a full training dataset of 20,000,000 samples, a small dataset of 50,000 samples repeated across 200 epochs, and an extremely small dataset of 10,000 samples repeated across 200 epochs. In totality, we can see that even though we reduced the dataset size significantly, the performance at 50,000 samples is still remarkable. The models retain the ranking in performance in this scenario. However, reducing this dataset further sees ELG become the top performing model. This is likely because ELG leverages distinct features in the form of polar coordinates in a small k-Nearest Neighborhood. By having these explicit features, the model requires less data to translate it to a meaningful representation. Whereas for our model, we learn the importance of the pairings entirely in the latent space, requiring more data. It should also be noted that our representation of locality is likely a superset of ELG's approach.

\begin{table*}[ht!]
\caption{Optimality gaps across various countries and dataset sizes. Best performing models in bold.}
\label{tbl:bar_plot}
\begin{tabular}{@{}ccccccc@{}}
\toprule \midrule
\multicolumn{1}{l}{Dataset}   & \# of Samples               & Model                         & Tour Length     & Opt. Gap (\%)                          & Aug. Tour Length & Aug. Opt. Gap (\%) \\ \midrule \midrule
\multicolumn{1}{c}{}         & \multicolumn{1}{c}{}       & \multicolumn{1}{c}{Concorde} & 5.6209          & \multicolumn{1}{c}{-}                 & 5.6209           & -                  \\
\multicolumn{1}{c}{}         & \multicolumn{1}{c}{}       & \multicolumn{1}{c}{POMO}     & 5.7190          & \multicolumn{1}{c}{1.7458\%}          & 5.7166           & 1.7024\%           \\
\multicolumn{1}{c}{USA13509} & \multicolumn{1}{c}{50,000} & \multicolumn{1}{c}{Sym-NCO}  & 5.7345          & \multicolumn{1}{c}{2.0394\%}          & 5.6912           & 1.2691\%           \\
\multicolumn{1}{c}{}         & \multicolumn{1}{c}{}       & \multicolumn{1}{c}{ELG}      & 5.6848          & \multicolumn{1}{c}{1.1361\%}          & 5.6824           & 1.0936\%           \\
\multicolumn{1}{c}{}         & \multicolumn{1}{c}{}       & \multicolumn{1}{c}{Ours}     & \textbf{5.6769} & \multicolumn{1}{c}{\textbf{0.9954\%}} & \textbf{5.6750}  & \textbf{0.9617\%}  \\ \midrule
\multicolumn{1}{c}{}         & \multicolumn{1}{c}{}       & \multicolumn{1}{c}{Concorde} & 5.6209          & \multicolumn{1}{c}{-}                 & 5.6209           & -                  \\
\multicolumn{1}{c}{}         & \multicolumn{1}{c}{}       & \multicolumn{1}{c}{POMO}     & 5.8875          & \multicolumn{1}{c}{4.7433\%}          & 5.8848           & 4.6950\%           \\
\multicolumn{1}{c}{USA13509} & \multicolumn{1}{c}{10,000} & \multicolumn{1}{c}{Sym-NCO}  & 5.8598          & \multicolumn{1}{c}{4.2695\%}          & 5.8125           & 3.4282\%           \\
\multicolumn{1}{c}{}         & \multicolumn{1}{c}{}       & \multicolumn{1}{c}{ELG}      & \textbf{5.7849} & \multicolumn{1}{c}{\textbf{2.9172\%}} & \textbf{5.7771}  & \textbf{2.7782\%}  \\
\multicolumn{1}{c}{}         & \multicolumn{1}{c}{}       & \multicolumn{1}{c}{Ours}     & 5.8152          & \multicolumn{1}{c}{3.4575\%}          & 5.8079           & 3.3265\%           \\ \midrule \midrule
\multicolumn{1}{c}{}         & \multicolumn{1}{c}{}       & \multicolumn{1}{c}{Concorde} & 3.5341          & \multicolumn{1}{c}{-}                 & 3.5341           & -                  \\
\multicolumn{1}{c}{}         & \multicolumn{1}{c}{}       & \multicolumn{1}{c}{POMO}     & 3.5699          & \multicolumn{1}{c}{1.0141\%}          & 3.5697           & 1.0073\%           \\
\multicolumn{1}{c}{JA9847}   & \multicolumn{1}{c}{50,000} & \multicolumn{1}{c}{Sym-NCO}  & 3.5724          & \multicolumn{1}{c}{1.0837\%}          & 3.5622           & 0.7960\%           \\
\multicolumn{1}{c}{}         & \multicolumn{1}{c}{}       & \multicolumn{1}{c}{ELG}      & 3.5508          & \multicolumn{1}{c}{0.4723\%}          & \textbf{3.5504}  & \textbf{0.4627\%}  \\
\multicolumn{1}{c}{}         & \multicolumn{1}{c}{}       & \multicolumn{1}{c}{Ours}     & \textbf{3.5506} & \multicolumn{1}{c}{\textbf{0.4667\%}} & 3.5504           & 0.4629\%           \\ \midrule 
\multicolumn{1}{c}{}         & \multicolumn{1}{c}{}       & \multicolumn{1}{c}{Concorde} & 3.5341          & \multicolumn{1}{c}{-}                 & 3.5341           & -                  \\
\multicolumn{1}{c}{}         & \multicolumn{1}{c}{}       & \multicolumn{1}{c}{POMO}     & 3.6137          & \multicolumn{1}{c}{2.2514\%}          & 3.6136           & 2.2494\%           \\
\multicolumn{1}{c}{JA9847}   & \multicolumn{1}{c}{10,000} & \multicolumn{1}{c}{Sym-NCO}  & 3.6037          & \multicolumn{1}{c}{1.9695\%}          & \textbf{3.5841}  & \textbf{1.4154\%}  \\
\multicolumn{1}{c}{}         & \multicolumn{1}{c}{}       & \multicolumn{1}{c}{ELG}      & \textbf{3.5868} & \multicolumn{1}{c}{\textbf{1.4917\%}} & 3.5863           & 1.4779\%           \\
\multicolumn{1}{c}{}         & \multicolumn{1}{c}{}       & \multicolumn{1}{c}{Ours}     & 3.5943          & \multicolumn{1}{c}{1.7037\%}          & 3.5942           & 1.7016\%           \\ \midrule \midrule
\multicolumn{1}{c}{}         & \multicolumn{1}{c}{}       & \multicolumn{1}{c}{Concorde} & 5.0019          & \multicolumn{1}{c}{-}                 & 5.0019           & -                  \\
\multicolumn{1}{c}{}         & \multicolumn{1}{c}{}       & \multicolumn{1}{c}{POMO}     & 5.1228          & \multicolumn{1}{c}{2.4158\%}          & 5.1164           & 2.2883\%           \\
\multicolumn{1}{c}{BM33708}  & \multicolumn{1}{c}{50,000} & \multicolumn{1}{c}{Sym-NCO}  & 5.0878          & \multicolumn{1}{c}{1.7164\%}          & 5.0532           & 1.0242\%           \\
\multicolumn{1}{c}{}         & \multicolumn{1}{c}{}       & \multicolumn{1}{c}{ELG}      & 5.0648          & \multicolumn{1}{c}{1.2570\%}          & 5.0572           & 1.1055\%           \\
\multicolumn{1}{c}{}         & \multicolumn{1}{c}{}       & \multicolumn{1}{c}{Ours}     & \textbf{5.0566} & \multicolumn{1}{c}{\textbf{1.0925\%}} & \textbf{5.0520}  & \textbf{0.9833\%}  \\ \midrule
\multicolumn{1}{c}{}         & \multicolumn{1}{c}{}       & \multicolumn{1}{c}{Concorde} & 5.0019          & \multicolumn{1}{c}{-}                 & 5.0019           & -                  \\
\multicolumn{1}{c}{}         & \multicolumn{1}{c}{}       & \multicolumn{1}{c}{POMO}     & 5.1732          & \multicolumn{1}{c}{3.4248\%}          & 5.1659           & 3.2784\%           \\
\multicolumn{1}{c}{BM33708}  & \multicolumn{1}{c}{10,000} & \multicolumn{1}{c}{Sym-NCO}  & 5.1666          & \multicolumn{1}{c}{3.2914\%}          & 5.1307           & 2.5740\%           \\
\multicolumn{1}{c}{}         & \multicolumn{1}{c}{}       & \multicolumn{1}{c}{ELG}      & \textbf{5.1173} & \multicolumn{1}{c}{\textbf{2.3064\%}} & \textbf{5.1080}  & \textbf{2.2109\%}  \\
\multicolumn{1}{c}{}         & \multicolumn{1}{c}{}       & Ours                          & 5.5157          & 2.9934\%                               & 5.1440           & 2.8393\%           \\ \midrule \bottomrule 
\end{tabular}
\end{table*}

\subsection{Comparison with General Solvers}
\begin{table*}[ht!]
\caption{Comparison of GLOP and our model on the various datasets.}
\label{tbl:glop}
\begin{tabular}{@{}ccccc@{}}
\toprule \midrule
Dataset  & Model                         & Aug. Tour Length & Aug. Opt. Gap (\%) & Run-time              \\ \midrule
         & \multicolumn{1}{c|}{Concorde} & 5.6209           & -                  &                       \\
USA13509 & \multicolumn{1}{c|}{GLOP}     & 5.6704           & 0.8799\%           & 6 min 07 sec          \\
         & \multicolumn{1}{c|}{Ours}     & \textbf{5.6533}  & \textbf{0.5762\%}  & \textbf{3 min 03 sec} \\ \midrule
         & \multicolumn{1}{c|}{Concorde} & 3.5341           & -                  &                       \\
JA9847   & \multicolumn{1}{c|}{GLOP}     & 3.5839           & 1.4094\%           & 6 min 04 sec          \\
         & \multicolumn{1}{c|}{Ours}     & \textbf{3.5435}  & \textbf{0.2670\%}  & \textbf{3 min 01 sec} \\ \midrule
         & \multicolumn{1}{c|}{Concorde} & 5.0019           & -                  &                       \\
BM33408  & \multicolumn{1}{c|}{GLOP}     & 5.0510           & 0.9822\%           & 6 min 07 sec          \\
         & \multicolumn{1}{c|}{Ours}     & \textbf{5.0328}  & \textbf{0.6166\%}  & \textbf{3 min 04 sec} \\ \bottomrule
\end{tabular}
\end{table*}

While our approach biases the solver towards the distribution, recent works such as GLOP \cite{ye2024glop} proposed generic solvers that utilize the attention model as a Hamiltonian path solver. Table \ref{tbl:glop} compares our neural solver against GLOP. As GLOP's main premise is to break down a large problem and solve multiple sub-paths using local solvers trained on smaller sizes. Since our test results are on TSP100, we utilize the TSP25 and TSP50 solvers. Overall, we can see that our approach is stronger than a generically trained solver in both speed and performance. Nevertheless, since GLOP is based on the transformer model, our contributions can easily be integrated.

\end{document}